\documentclass[journal]{IEEEtran}
\usepackage[colorlinks,urlcolor=blue,linkcolor=blue,citecolor=blue]{hyperref}
\usepackage{wrapfig}
\usepackage{graphicx}
\usepackage{titlesec}
\usepackage{bbding}

\def\eg{\emph{e.g.\ }} 
\def\ie{\emph{i.e.\ }} 
 
\def\etc{\emph{etc.\ }} 
 
\def\etal{\emph{et al.\ }}
\usepackage{array}
\usepackage{booktabs}
\usepackage{algorithm}  
\usepackage{algorithmic}
\usepackage{amssymb}
\usepackage{amsmath}
\usepackage{amssymb}
\usepackage{makecell}
\newcommand{\tabincell}[2]{\begin{tabular}{@{}#1@{}}#2\end{tabular}}  
\usepackage{bm}
\usepackage{multirow}
\usepackage{svg} 
\usepackage{amsmath,amsthm,amssymb,amsfonts}

\ifCLASSINFOpdf
\else
\fi
\begin{document}
%
\title{Low Rank Optimization for Efficient Deep Learning: Making A Balance between Compact Architecture and Fast Training}
%
%
%

\author{Xinwei~Ou, Zhangxin~Chen, Ce~Zhu,~\IEEEmembership{Fellow,~IEEE}, and~Yipeng~Liu,~\IEEEmembership{Senior Member,~IEEE}
\thanks{All the authors are with the School of Information and Communication
Engineering, University of Electronic Science and Technology of China (UESTC), Chengdu 611731 China (e-mails: xinweiou@std.uestc.edu.cn, $ \{$zhangxinchen, eczhu, yipengliu$\}$@uestc.edu.cn).
}

\thanks{Manuscript received ×× ××, 2023; revised ×× ××, 2023.}}

\maketitle

\begin{abstract}
Deep neural networks have achieved great success in many data processing applications. However, the high computational complexity and storage cost makes deep learning hard to be used on resource-constrained devices, and it is not environmental-friendly with much power cost. In this paper, we focus on low-rank optimization for efficient deep learning techniques. In the space domain, deep neural networks are compressed by low rank approximation of the network parameters, which directly reduces the storage requirement with a smaller number of network parameters. In the time domain, the network parameters can be trained in a few subspaces, which enables efficient training for fast convergence. The model compression in the spatial domain is summarized into three categories as pre-train, pre-set, and compression-aware methods, respectively. With a series of integrable techniques discussed, such as sparse pruning, quantization, and entropy coding, we can ensemble them in an integration framework with lower computational complexity and storage. Besides of summary of recent technical advances, we have two findings for motivating future works: one is that the effective rank outperforms other sparse measures for network compression. The other is a spatial and temporal balance for tensorized neural networks.
\end{abstract}

\begin{IEEEkeywords}
model compression, subspace training, effective rank, low rank tensor optimization, efficient deep learning.
\end{IEEEkeywords}

%
\IEEEpeerreviewmaketitle

\section{Introduction}
%
%
%
%
\IEEEPARstart{D}{eep} neural networks (DNNs) have been widely used in many data processing applications, such as speech recognition, computer vision \cite{krizhevsky2012imagenet, simonyan2014very, jiang2017exploiting}, natural language processing \cite{vaswani2017attention, graves2013speech}, \etc As a deeper or wider structure can lead to better performance, DNNs are gradually characterized by their over-parameterization. Over-parameterization, on the other hand, suggests too much redundancy in DNNs, which leads to overfitting \cite{hinton2012improving, denil2013predicting}. There are mainly two challenges in
deep learning: high complexity and slow convergence. The high complexity means that there are millions of parameters in DNNs, and computation between massive parameters and inputs is cumbersome, which underlines the need for efficient algorithms to compress and accelerate. For example, the number of parameters in VGG-16 \cite{simonyan2014very} is almost seven million. For an image in ImageNet dataset \cite{krizhevsky2012imagenet} with a size of 224×224×3, the feedforward process requires 30.9 billion float point-operations (FLOPs). The high complexity is unaffordable for resource-limited devices, such as mobile phones \cite{kim2015compression} and IoT devices \cite{lane2015early}. The slow convergence is caused by the conventional back propagation (BP) algorithm, resulting in time-consuming training \cite{abdul2011accelerating}. Also, the convergence speed is sensitive to the setting of the learning rate and the way to initialize weights.

\par There are many works attempting to reduce the high complexity of DNNs with acceptable performance decay. The investigation of model compression can be summarized into two folds: one is to reduce the number of parameters, and the other is to reduce the average bit width of data representation. The first fold includes but is not limited to low rank approximation \cite{lebedev2014speeding,jaderberg2014speeding,wang2018wide, liu2022tensor}, pruning \cite{luo2017thinet, zhang2018systematic}, weight-sharing \cite{ullrich2017soft}, sparsity \cite{huang2011learning}, knowledge distillation \cite{han2015deep}. Since these techniques have their own limitations, it is better to combine them to exploit the redundancy in DNNs fully. Quantization \cite{gong2014compressing, wu2016quantized} and entropy coding \cite{han2015deep} belong to the second category, which is designed to achieve a lower number of bits per parameter. 
\par Low rank approximation has been widely adopted due to its strong theoretical basis and ease of implementation on hardware. In this survey, we comprehensively review this rapidly developing area by dividing low rank optimization for model compression into three main categories: pre-train method, pre-set method, and compression-aware method. The biggest distinction among them is the way to train. The pre-train method directly decomposes a pre-trained model to get warm initialization for the compressed format, followed by retraining the compressed model to recover the performance. Without pre-training, the pre-set method trains a network that is pre-set to a compact format from scratch. Totally different from the above two methods, the compression-aware method explicitly accounts for compression in the training process by gradually enforcing the network to enjoy low-rank structure. Although the discussion about low rank optimization can also be found in \cite{wang2023tensor}, we further investigated how to integrate it with other compression techniques to pursue lower complexity and recommended the effective rank as the most efficient measure used in low rank optimization. 

\par When the redundancy in DNNs is exploited by subspace training, DNNs can converge faster without losing accuracy. In deep learning, it is conventional to train networks with first-order optimization methods, \eg SGD \cite{ruder2016overview}, which is computationally cheap. But there are some inherent drawbacks to first-order optimization methods, such as slow theoretical and empirical convergence. Second-order methods can deal with such a problem well, but because of the heavy computational burden of Hessian matrices, second-order methods are not applicable to DNNs. The idea that projecting parameters onto a tiny subspace represented by several independent variables is an effective way to solve this problem. Since only a few variables need to be optimized, we can apply second-order optimization methods to achieve the temporal efficiency of deep learning.
\begin{figure*}[!htbp]
\begin{center}
 \centering
    \includegraphics[scale=0.7]{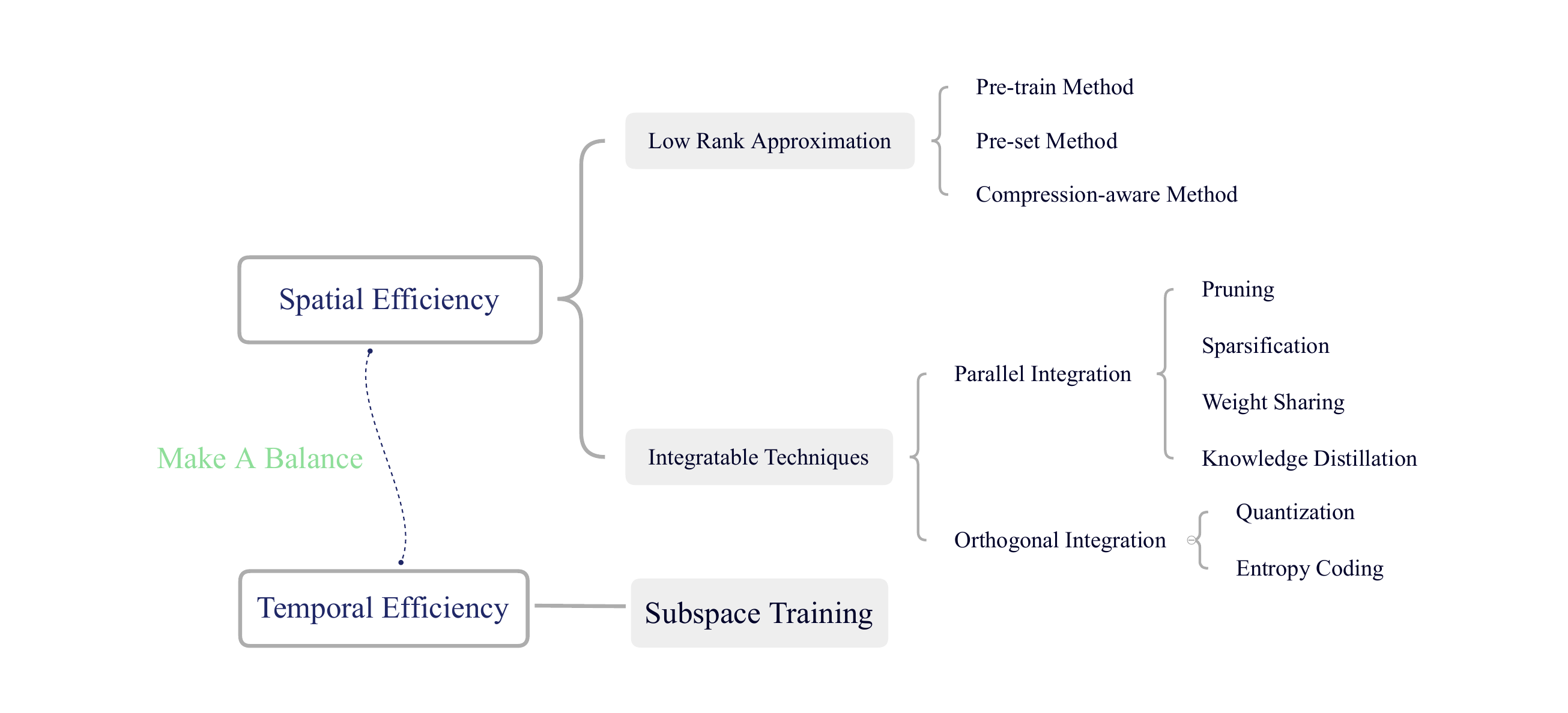}
    \caption{Overview of low rank tensor optimization for efficient deep learning.}
    \label{overview}
\end{center}
\end{figure*}
\par In this survey, we first present a comprehensive overview of various tensor decomposition methods applicable to model compression. Next, the low rank optimization for model compression is summarized in terms of pre-set, pre-train, and compression-aware methods. For each method, a detailed discussion on key points about implementation is given. More meticulously, we make a comparison among various sparsity measures used in the compression-aware method, and dig out the most efficient measure, \ie{effective rank}, which is seldom used as a sparse regularizer before. In addition, while there are already many works that give a list of joint-way compression \cite{deng2020model, choudhary2020comprehensive}, little attention has been paid to the integration between low rank approximation and other compression techniques. Therefore, we present an overall survey on this kind of integration here. Then, we introduce low rank optimization for subspace training. Furthermore, we are the first to relate these two types of low rank optimization, discovering that redundancy in the temporal domain and spatial domain are of the same origin. And there is a discussion on how to apply subspace training on tensorized neural networks to achieve spatial efficiency and temporal efficiency simultaneously.
The overall mind map to achieve efficient deep learning through low rank optimization is shown in Fig. \ref{overview}.

\par Different from the previous surveys on tensors for efficient deep learning \cite{liu2022tensor, liu2021tensor, liu2021tensors}, the main contributions of this paper can be summarized as follows.
\begin{itemize}
  \item [1)] 
  We make a detailed overview of low rank approximation for model compression, and we find that RNNs can be effectively compressed using Hierarchical Tucker (HT) decomposition and Kronecker Product Decomposition (KPD), CNNs can be effectively compressed using Tensor Train (TT) and Generalized Kronecker Product Decomposition (GKPD), while Tensor Ring (TR) and Block Term Decomposition (BTD) can suitably compress both RNNs and CNNs.

    \item [2)]
  A series of integratable neural network compression techniques have been discussed in details, and an integration framework is summarized to well take advantage of various methods. 
  \item [3)]
  We analyse that the redundancy in the space domain and time domain are of the same origin. In order to accelerate the training of tensorized neural networks, we should make the balance between spatial efficiency and temporal efficiency.
  \item [4)]
  After discussing and testing various sparse measures for low rank optimization for deep neural network compression, the effective rank outperforms in numerical experiments.
\end{itemize}

\par This survey is organized as follows. In Section \ref{sec:2}, we give an overview of low rank optimization for model compression. Low rank approximation integrated with other compression techniques is reviewed in Section \ref{sec:3}. Section \ref{sec:4} introduces low rank optimization for subspace training and analyses the coupling between these two types of low rank optimization.

 

\section{Low Rank Optimization for Model Compression} \label{sec:2}

Since DNNs are over-parameterized, there are  opportunities to make deep networks more compact. Compression methods, like quantization, pruning and low-rank approximation, can lower complexity of DNNs without significant accuracy degradation. Among them, low rank approximation has been widely adopted because of the solid theoretical basis of tensor decomposition. In this section, we will first introduce various tensor decomposition methods applicable for network compression, and then divide optimization methods for low rank approximation into three categories: pre-train, pre-set, and compression-aware methods. In addition, we make a discussion on efficient sparsity measures.

\subsection{Tensor Decomposition}
\par Low rank approximation can provide an ultra-high compression ratio for recurrent neural network (RNN) with insignificant accuracy loss. However, when it comes to convolutional neural network (CNN), the compression performance isn't as satisfying as RNN. In early literatures, 4D convolutional kernels are reshaped into matrices and singular value decomposition (SVD) is utilized to decompose matrices into two factors \cite{zhang2015accelerating}. However, the reshaping operation leads to distortion of structure information. Hence, more efficient tensor decomposition has attracted interests. Canonical-Polyadic (CP) decomposition \cite{liu2022tensor} is applied to decompose a convolutional layer into four consecutive convolutional layers, significantly speeding up CNNs \cite{lebedev2014speeding}. Tucker decomposition \cite{tucker1963implications} can decompose the 4D kernel into a 4D compact kernel and two matrices by exploiting the channel-wise redundancy. Based on these three classic decompositions, there are many other flexible methods emerged including HT \cite{grasedyck2010hierarchical}, TT \cite{oseledets2011tensor}, TR \cite{zhao2016tensor}, BTD  \cite{de2008decompositions}, GKPD \cite{hameed2022convolutional}, Semi-tensor Product (STP)-based Tensor Decomposition \cite{zhao2021semi}, which dramatically improve the compression performance for DNNs. Table \ref{tab:methods} shows the performance of widely-used tensor decomposition methods applied to compress ResNet32 with Cifar10 dataset.

\begin{table}[!htbp] 
\caption{Comparison of compression performance of advanced tensor decomposition methods on ResNet32 with Cifar10 dataset}
\label{tab:methods}
\centering
\begin{tabular}{lcc}
\toprule
Method & Top-1 Accuracy (\%) & Compression Ratio   \\
\midrule
Tucker \cite{kim2015compression} & 87.70 & 5× \\
TT \cite{garipov2016ultimate} & 88.3 & 4.8× \\
TR \cite{wang2018wide} & 90.6 & 5×  \\
BTD \cite{ye2020block} & 91.1 & 5×  \\
GKPD \cite{hameed2022convolutional} & 91.5 & 5×   \\
HT \cite{wu2020hybrid} & 89.9 & 1.6× \\
STT \cite{zhao2021semi} & 91.0 & 9× \\
\bottomrule
\end{tabular}
\end{table}

\par Here, we outline some key notations. For a fully-connected (FC) layer, we let $\mathbf{W}\in \mathbb{R}^{O\times I}$ denote the weight matrix of this layer, where $I$ and $O$ represent the number of input neurons and output neurons, respectively. And for a convolutional (Conv) layer, we let $\mathbf{\mathcal{K}} \in \mathbb{R}^{S\times C \times H \times W}$ denote the weight of the convolutional kernel, where $S$, $C$ are the number of filters and input channels, $H$, $W$ are the height and width of the kernel. In some cases, we need to reshape a tensor into a higher-order one. We assume that $I_1 \times I_2 \times \cdots \times I_d = I$, $O_1 \times O_2 \times \cdots \times O_d = O$, $C_1 \times C_2 \times \cdots \times C_d = C$,  and $S_1 \times S_2 \times \cdots \times S_d = S$. Some necessary mathematical operators are listed in Table \ref{notation}.
\begin{table}[!htbp]
     \caption{Notations used in this paper.}
    \label{notation}
    \centering
    \begin{tabular}{c|c}
    \hline
    Notations &  Descriptions \\
    \hline
    diag() &  \tabincell{c}{generate a diagonal matrix by taking \\ the input vector as the main diagonal}  \\
    \hline
    $\otimes$ & Kronecker product \\
    \hline
    $\circ$ & vector outer product \\
    \hline
    $\times_n$ & n-mode product \\
    \hline
    $\ltimes$ & semi-tensor product\\
    \hline
    
    \end{tabular}
   
\end{table}

\par Base on these defined notation, we can make a comparison among various state-of-art tensor decompositions on their ability to compress and accelerate. When aiming at FC layers, the comparison is shown in Table \ref{fc}. And Table  \ref{conv} is for Conv layers.
\begin{table}[!htbp]
    \centering
    \caption{Comparison among fully-connected layer compressed by TT, TR, HT, BTD, STR, and KPD on computation costs and storage consumption. Note that $I_m = \mathop{\max}_{k\in \{1,\cdots,d\}} I_k$, $O_m = \mathop{\max}_{k\in \{1,\cdots,d\}}O_k$, $d=2$ for KPD, $r$ is the maximal rank, $R$ is the CP rank of BTD, and $t$ is the ratio between connected dimensionality.}
    \label{fc}
    \footnotesize
    \begin{tabular}{c|l|l}
    \hline
     Method  & Computation & Storage  \\
     \hline
     FC & $\mathcal{O}(IO)$ & $\mathcal{O}(IO)$\\
     TT & $\mathcal{O}(dI_m\mathop{\max}(I,O)r^2)$ & $\mathcal{O}(dI_mO_mr^2)$\\
     TR & $\mathcal{O}(d(I+O)r^3)$ & $\mathcal{O}(d(I_m+O_m)r^2)$\\
     HT & $\mathcal{O}(d\mathop{\min}(I,O)(r^3+I_mr^2))$ & $\mathcal{O}(dI_mO_mr+dr^3)$\\
     BTD  &  $\mathcal{O}(dI_m\mathop{\max}(I,O)r^dR)$ & $\mathcal{O}((dI_mO_mr+r^d)R)$\\
     STR & $\mathcal{O}(d(I+O)r^3/t)$ & $\mathcal{O}(d(I_m+O_m)r^2/t^2)$\\
     KPD & $\mathcal{O}(IO_m+OI_m)$ & $\mathcal{O}(I_mO_m)$\\
     \hline
    \end{tabular}
\end{table}

\begin{table*}[!htbp]
    
    \caption{Comparison among convolutional layer compressed by TT, TR, HT, BTD, STR, GKPD on computation costs and storage consumption. Note that $C_m = \mathop{\max}_{k\in \{1,\cdots,d\}} C_k$, $S_m = \mathop{\max}_{k\in \{1,\cdots,d\}}S_k$, $d=2$ for GKPD, $k =  \mathop{\max}(k_1,k_2)$ with $k_1*k_2=K$, $r$ is the maximal rank, $R$ is the CP rank of BTD, $M$ and $N$ are the height and width of feature map, and $t$ is the ratio between connected dimensionality.}
    \label{conv}
    \centering
    \begin{center}
    \begin{tabular}{c|l|l}
    \hline
     Method  & Computation & Storage  \\
     \hline
     Conv & $\mathcal{O}(SCK^2MN)$ & $\mathcal{O}(SCK^2)$\\
     TT & $\mathcal{O}(dr\mathop{\max}(rC_m,K^2)\mathop{\max}(C,S)MN)$ & $\mathcal{O}(dC_mS_mr^2+K^2r)$\\
     TR & $\mathcal{O}(r^3(C+S)+(r^3K^2+r^2(C+S))MN)$ & $\mathcal{O}((dC_mS_m+K^2)r^2)$\\
     HT & $\mathcal{O}(\log_{2}d CS(r^3+r^2)+SCK^2MN)$ & $\mathcal{O}(dC_mS_mr+K^2r+dr^3)$\\
     BTD & $\mathcal{O}((K^2r^2+(C+S)r)RMN)$ & $\mathcal{O}((K^2r^2+(I+O)r)R)$\\
     STR & $\mathcal{O}(\frac{r^3}{t^3}(C+S)+(r^3K^2+\frac{r^2}{t}(C+S))MN)$ & $\mathcal{O}((\frac{dC_mS_m}{t^2}+K^2)r^2)$\\
     GKPD & $\mathcal{O}(r(C_mS+S_mC)k^2MN)$ & $\mathcal{O}(rC_mS_mk^2)$\\
     \hline
    \end{tabular}
    \end{center}
\end{table*}

\subsubsection{Singular Value Decomposition}
\par For a given matrix $\mathbf{X}\in \mathbb{R}^{M\times N}$, its SVD can be written as
\begin{equation}
\mathbf{X}=\mathbf{U}\operatorname{diag}(\mathbf{s})\mathbf{V}^\text{T}.
\end{equation}  
Let $R$ denote the rank of the matrix, $R\leq \min\{M,N\}$. Note that $\mathbf{U}\in \mathbb{R}^{M\times N}$ and $\mathbf{V}\in \mathbb{R}^{N\times R}$ satisfy $\mathbf{U}\mathbf{U}^T=\mathbf{I}$ and $\mathbf{V}\mathbf{V}^T=\mathbf{I}$, respectively. And the elements of $\mathbf{s} \in \mathbb{R}^R$ decrease from first to end, \ie{$s_1 \geq s_2 \geq .. \geq s_R$}.
\par Since the format of weights in FC layers is a natural matrix, SVD can be directly utilized. By using SVD, the FC layer can be approximated by two consecutive layers.
The weight of the first and second layer can be represented by $\mathbf{B}=diag(\mathbf{\sqrt{s}})\mathbf{V}^T$ and $\mathbf{A}=\mathbf{U}diag(\mathbf{\sqrt{s}})$, respectively. For Conv layers, the 4D kernel should be reshaped into a 2D matrix first. By exploiting different types of redundancy, there are two decomposition schemes, one reshapes $\mathbf{\mathcal{W}}$ into a $S$-by-$CHW$ matrix, namely channel-wise decomposition \cite{zhang2015accelerating}, the other called spatial-wise decomposition \cite{jaderberg2014speeding} gets a $SH$-by-$CW$ matrix. Then, compute SVD of the reshaped matrix. Similar to the process of compressing FC layers, two Conv layers represented by tensors reshaped from factors $\mathbf{B}$ and $\mathbf{A}$ can be used to replace the original layer.
\par However, both methods only can exploit one type of redundancy. Moreover, there is also redundancy between input channels. Exploiting all kinds of redundancy at the same time can help us achieve a much higher compression ratio, which can be achieved by tensor decomposition.

\subsubsection{CP Decomposition}
While SVD factorizes a matrix into a sum of rank-one matrices, CP decomposition factorizes a tensor into a sum of rank-one tensors. For a $N$th order tensor, $\mathbf{\mathcal{X}}\in \mathbb{R}^{I_1\times I_2\times \cdots \times I_N}$, the CP decomposition can be formulated as:
\begin{equation}
    \mathbf{\mathcal{X}} = \left[\kern-0.35em\left[ \mathbf{\lambda}; \mathbf{A}^{(1)},\mathbf{A}^{(2)}, \cdots, \mathbf{A}^{N} \right]\kern-0.35em\right] = 
    \sum_{r=1}^{R}\lambda_r \mathbf{a}^{(1)}_r \circ \mathbf{a}^{(2)}_r \circ \cdots \circ \mathbf{a}^{(N)}_r.
\end{equation}
 Each $\mathbf{a}_r^{(n)}$ represents the $r$th column of $\mathbf{A}^{(n)}$ and $\mathbf{\lambda}\in \mathbb{R}^{R}$ represents the significance of $R$ components. The rank of the tensor $\mathbf{\mathcal{X}}$, denoted by $R$, is defined as the smallest number of rank-one tensors \cite{liu2021tensor, liu2019low}.

\par When using CP to compress FC layers, the weight matrix $\mathbf{W}$ should be firstly tensorized into a $2d$th order tensor $\mathbf{\mathcal{W}} \in \mathbb{R}^{O_1\times O_2 \cdots \times O_d \times I_1 \times I_2 \times \cdots \times I_d} $. Meanwhile, the input vector $\mathbf{x}\in \mathbb{R}^{I}$ should be presented as a $d$th order tensor $\mathbf{\mathcal{X}}\in \mathbb{R}^{I_1 \times I_2 \times \cdots \times I_d}$. For convolutional kernels, by directly performing CP on the 4-D kernel tensor, the layer will be approximated by four consecutive convolutional layers whose weights are represented by four factor matrices, respectively.

\subsubsection{Tucker Decomposition}
The Tucker decomposition can be considered as a higher-order generalization of principal component analysis (PCA). It represents an $N$-th order tensor with a $N$th order core tensor multiplied by a basis matrix along each mode. Thus, for $\mathbf{\mathcal{X}} \in \mathbb{R}^{I_1 \times I_2 \times \cdots \times I_N}$, we have
\begin{equation}\label{tk}
    \mathbf{\mathcal{X}} = \mathbf{\mathcal{G}} \times_1 \mathbf{A}^{(1)} \times_2 \mathbf{A}^{(2)} \times_3 \cdots \times_N \mathbf{A}^{(N)},
\end{equation}
Where $\mathbf{\mathcal{G}}\in \mathbb{R}^{R_1\times R_2\times \cdots \times R_N}$ is called core tensor.
Here, "$\times_n$" represents the $n$-mode product, \ie{multiply a tensor by a matrix in mode n}. Elementwise, "$\times_n$" can be formulated as:
\begin{equation}
    (\mathbf{\mathcal{G}} \times_1 \mathbf{A}^{(1)})_{i_1,r_2,\cdots,r_N} = \sum_{r_1 = 1}^{R_1}
    \mathbf{\mathcal{G}}_{r_1, r_2, \cdots, r_N}\mathbf{A}^{(1)}_{i_1, r_1}.
\end{equation}
Columns of the factor matrix $\mathbf{A}^{(n)}\in \mathbb{R}^{I_n \times R_n}$ can be considered as the principal components of the $n$th mode. The core tensor $\mathbf{\mathcal{G}}$ can be viewed as a compressed version of $\mathbf{\mathcal{X}}$ or the coefficient in the low dimensional subspace. 
In this case, we can say that $\mathbf{\mathcal{X}}$ is a rank-($R_1, R_2, \cdots, R_N$) tensor \cite{liu2021tensor, liu2019low}. 

 \par In the case of compressing FC layers, similar to CP, the same tensorization for weights and input is needed, followed by directly performing Tucker decomposition on the $2d$th order tensor. For Conv layers, since the spatial size of the kernel is too small, we can just use Tucker2 \cite{tucker1966some} to take advantage of redundancy between filters and between input channels, generating 1$\times$1 convolution, $H\times W$ convolution, and 1$\times$1 convolution.
 
 \subsubsection{Block Term Decomposition}
 Block Term Decomposition (BTD) was introduced in \cite{de2008decompositions} as a more powerful tensor decomposition, which combines the CP decomposition and Tucker decomposition. Consequently, BTD is more robust than the original CP and Tucker decomposition. While CP approximates a tensor with a sum of rank-one tensors, BTD is a sum of tensors in low rank Tucker format. Or, by concatenating factor matrices in each mode and arranging all the core tensors of each subtensor into a block diagonal core tensor, BTD can be considered as an instance of Tucker. Hence, consider a $N$th order tensor, $\mathbf{\mathcal{X}}\in \mathbb{R}^{I_1 \times I_2 \times \cdots \times I_d}$, its BTD can be written as:
 \begin{equation}
     \mathbf{\mathcal{X}} = \sum_{n=1}^{N}{\mathbf{\mathcal{G}}_n \times_1 \mathbf{A}^{(1)}_{n} \times_2 \mathbf{A}^{(2)}_{n}} \times_3 
     \cdots \times_d \mathbf{A}^{(d)}_{n}.
 \end{equation}
 In this formula, $N$ denotes the CP rank, \ie{the number of block terms}, and $\mathbf{\mathcal{G}}_n \in \mathbb{R}^{R_1\times R_2 \times \cdots \times R_d}$ is the core tensor of the $n$-th block term with multilinear ranks that equals $(R_1, R_2, \cdots, R_d)$.
 \par When BTD is applied to compress an FC layer, the yielded compact layer is called Block Term Layer (BTL) \cite{ye2020block}. In the BTL, the input tensor $\mathbf{\mathcal{X}}\in \mathbb{R}^{I_1\times I_2\times \cdots\times I_d}$ is tensorized from the original input vector $\mathbf{x} \in \mathbb{R}^{I}$ and the original weight matrix $\mathbf{W}$ is reshaped as $\mathbf{\mathcal{W}} \in \mathbb{R}^{O_1\times I_1 \times O_2 \times I_2 \times \cdots \times O_d \times I_d }$. Then, we can factorize $\mathbf{\mathcal{W}}$ by BTD with factor tensors $\{A^{(d)}_{n}\in \mathbb{R}^{O_d\times I_d\times R_d}\}_{n=1}^{d}$.
By conducting a tensor contraction operator between BTD($\mathbf{\mathcal{W}}$) and $\mathbf{\mathcal{X}}$, the output tensor $\mathbf{\mathcal{Y}}\in \mathbb{R}^{O_1\times O_2\cdots\times O_d}$ comes out, which can be vectorized as the final output vector.
For Conv layers, it is claimed in \cite{ye2020block} that by reshaping the 4D kernel into a matrix, $\mathbf{W}\in \mathbb{R}^{S\times CHW}$, the layer can be transformed into BTL. Specifically speaking, the matrix should be further reshaped as $1\times H\times 1\times W \times S_1 \times C_1 \times S_2 \times C_2 \times \cdots \times S_d \times C_d$.

\subsubsection{Hierarchical Tucker Decomposition}
\begin{figure}
    \centering
    \includegraphics[scale=0.5]{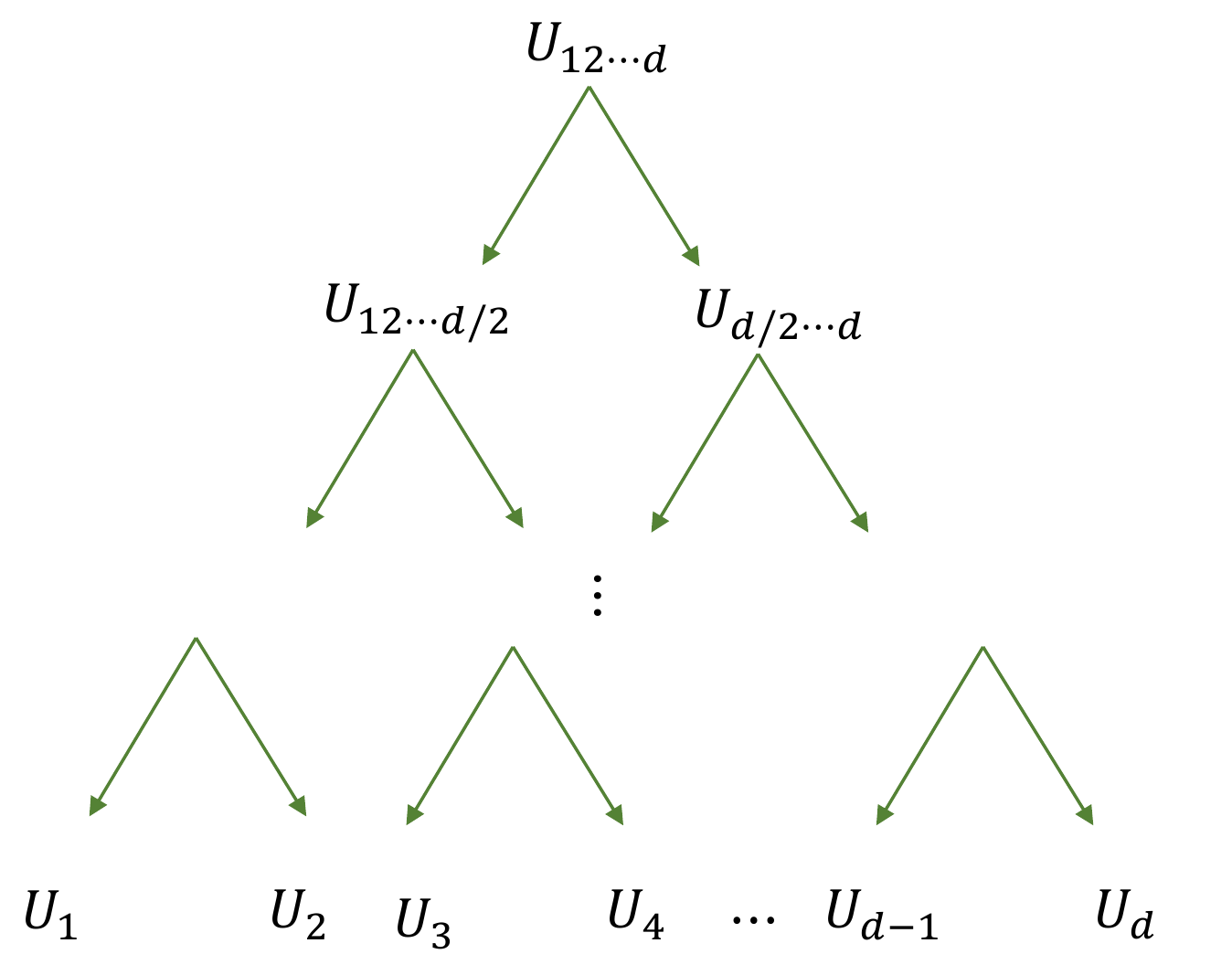}
    \caption{HT decomposition}
    \label{HT}
\end{figure}

Hierarchical Tucker (HT) decomposition is a hierarchical variant of the Tucker decomposition, which iteratively represents a high-order tensor with two lower-order subtensors and a transfer matrix via taking advantage of Tucker decomposition \cite{grasedyck2010hierarchical, liu2018image}.
For a tensor $\mathbf{\mathcal{X}}\in\mathbb{R}^{I_1\times I_2\times \cdots\times I_N}$, we can simply divide the index set $\{1,2,\cdots,N\}$ into two subsets, \ie{$\mathbb{T}=\{t_1, t_2, \cdots, t_k\}, \mathbb{S}=\{s_1, s_2, \cdots, s_{N-k}\}$}. Let $\mathbf{U}_{12\cdots N}\in \mathbb{R}^{I_{t_1} I_{t_2} \cdots I_{t_k} I_{s_1} I_{s_2} \cdots I_{s_{N-k}}\times 1}$ denotes the matrix reshaped from $\mathbf{\mathcal{X}}$, and truncated matrices $\mathbf{U}_{t}\in\mathbb{R}^{I_{t_1}I_{t_2}\cdots I_{t_k}\times R_t}$, $\mathbf{U}_{s}\in\mathbb{R}^{I_{s_1}I_{s_2}\cdots I_{s_{N-k}}\times R_s}$ represent the corresponding column basis matrix of two subspaces. Then, we could have:
\begin{equation}
    \mathbf{U}_{12\cdots N}=(\mathbf{U}_t\otimes \mathbf{U}_s)\mathbf{B}_{12\cdots N},
\end{equation}
where $\mathbf{B}_{12\cdots N} \in \mathbb{R}^{R_t R_s \times 1}$ is termed as transfer matrix and “$\otimes$” denotes the Kronecker product between two matrices. Subsequently, divide the set $\mathbb{T}$ into two subsets $ \mathbb{L}={\{l_1, l_2, \cdots, l_q\}}$ and $\mathbb{V}=\{v_1, v_2, \cdots, v_{k-q}\}$. We can represent $\mathbf{U}_t$ with $\mathbf{U}_l\in \mathbb{R}^{I_{l_1}I_{l_2}\cdots I_{l_{q}}\times R_l}$, $\mathbf{U}_v\in \mathbb{R}^{I_{v_1}I_{v_2}\cdots I_{v_{k-q}}\times R_v}$, and $\mathbf{B}_t\in \mathbb{R}^{R_l R_v \times R_t}$ as:
\begin{equation} \label{kro}
    \mathbf{U}_{t}=(\mathbf{U}_l\otimes \mathbf{U}_v)\mathbf{B}_{t}.
\end{equation}
The similar factorization procedure applies simultaneously to $\mathbf{U}_s$. By repeating this procedure until the index set cannot be divided, we can eventually obtain the treelike HT format of the target tensor. An illustration of a simple version of HT can be seen in Fig. \ref{HT}.

\par Since the Kronecker product in Eq. (\ref{kro}) is computationally expensive, there are other concise forms of HT, such as the contracted form introduced in \cite{wu2020hybrid}. This form merges index subsets to the universal set from bottom to top. 
In this form, an external input can  be contracted with each transfer matrix and truncated matrix one by one. This way can avoid the memory and computation-consuming weight reconstruction procedure and intermediate outputs will not be too large to out of memory. 
\begin{figure}
    \centering
    \includegraphics[scale=0.5]{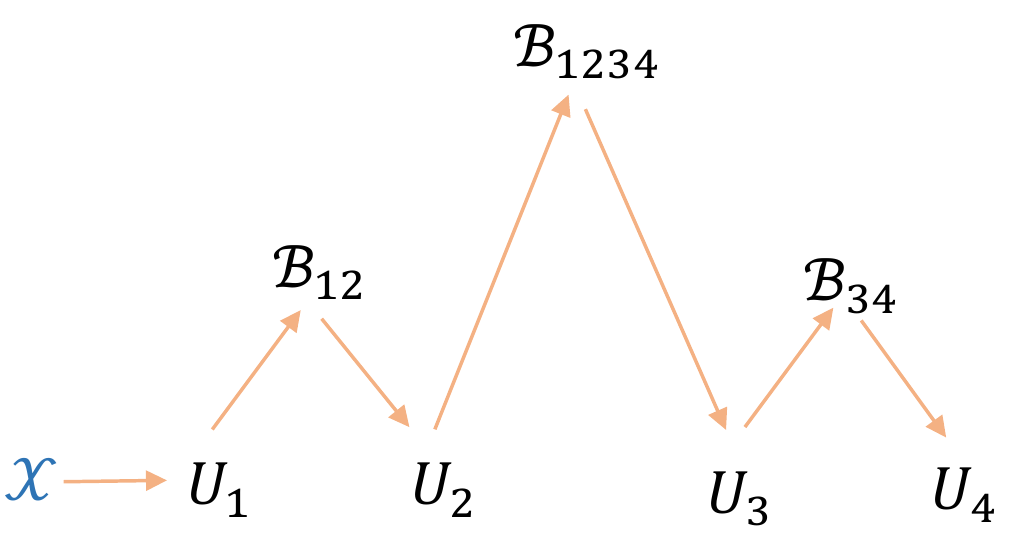}
    \caption{The chain computation for a 4th order case. $\mathcal{X}$ represents a 4th order input. These arrows represent the order of contraction.}
    \label{contract}
\end{figure}

\par For the realization of compressing FC layers by HT, the weight matrix should be transformed into $\mathbf{\mathcal{W}}\in \mathbb{R}^{(I_1 \cdot O_1)\times (I_2 \cdot O_2) \times \cdots \times (I_d \cdot O_d)}$, and the input data is tensorized into $\mathbf{\mathcal{X}}\in \mathbb{R}^{I_1 \times I_2 \times \cdots \times I_d}$. For reducing computation complexity, the chain computation shown in Fig. \ref{contract} is applied. However, as there is no law associating convolution and contraction, the kernel of Conv layers must be recovered from the HT format. By the way, in order to keep balance, the 4D kernel should be tensorized into $\mathbf{\mathcal{W}}\in \mathbb{R}^{(H\cdot W)\times (C_1 \cdot S_1) \times (C_2 \cdot S_2) \times \cdots \times (C_d \cdot S_d)}$.

\subsubsection{Tensor Train Decomposition}
\begin{figure}
    \centering
    \includegraphics[scale=0.5]{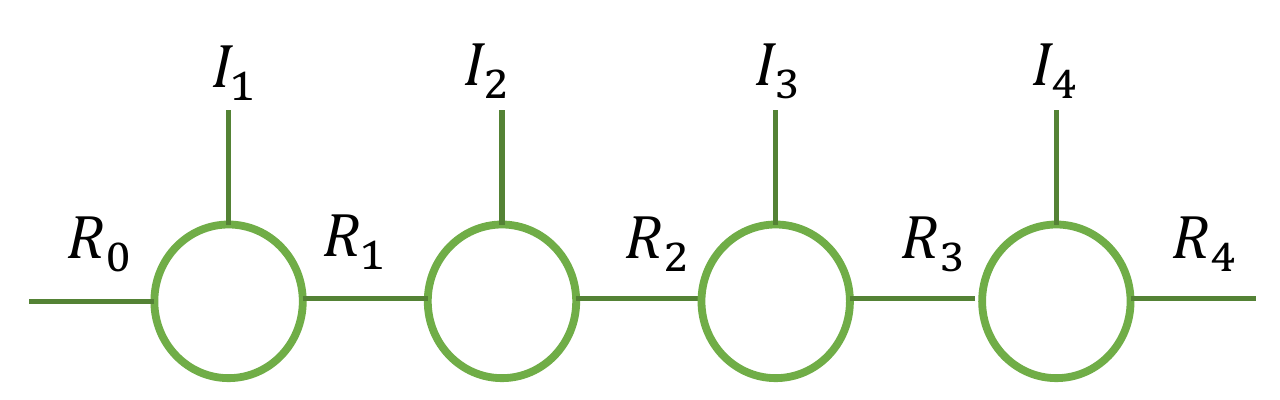}
    \caption{A 4th order tensor in TT format.}
    \label{TT}
\end{figure}

Tensor Train (TT) is a special case of HT, which is a degenerate HT format \cite{oseledets2011tensor, liu2020low}. TT factorizes a high-order tensor into a collection of 3rd or 2nd-order tensors. These core tensors are connected by the contraction operator. Assume that we have a $N$th order tensor, $\mathbf{\mathcal{X}}\in \mathbb{R}^{I_1\times I_2 \times \cdots\times I_N}$, elementwise, we can factorize it into TT format as:
\begin{equation}
\mathbf{\mathcal{X}}_{i_1,i_2,\cdots, i_N} = \sum_{r_1,r_2,\cdots,r_N} \mathbf{\mathcal{G}}^1_{i_1, r_1}\mathbf{\mathcal{G}}^2_{r_1,i_2, r_2}\cdots\mathbf{\mathcal{G}}^N_{r_{N-1},i_{N}},
\end{equation}
where the collection of $\{\mathbf{\mathcal{G}}^n \in \mathbb{R}^{R_{n-1}\times I_{n} \times R_{n}}\}_{n=1}^N$ with $R_0=1$ and $R_N=1$ is called TT-cores \cite{oseledets2011tensor}. The collection of ranks $\{R_n\}_{n=0}^{N}$ is called TT-ranks. Fig. \ref{TT} gives an illustration of a 4th order tensor represented in TT format.

\par The TT was first applied to compress FC layers in \cite{novikov2015tensorizing}, where the weight matrix is reshaped into a high order tensor, $\mathbf{\mathcal{W}}\in \mathbb{R}^{(I_1\cdot O_1)\times (I_2 \cdot O_2)\times \cdots \times (I_d \cdot O_d)}$. 
After representing $\mathbf{\mathcal{W}}$ in TT format, the resulted TT-cores $\{\mathbf{\mathcal{G}}^n \in \mathbb{R}^{R_{n-1}\times I_{n} \times O_{n}\times R_{n}}\}_{n=1}^N$ can directly be contracted with the tensorized input. It was suggested in \cite{wu2020hybrid} that TT is more efficient for compressing Conv layers than HT, while HT is more suitable for compressing FC layers whose weight matrix is more prone to be reshaped into a balanced tensor.
\par Employing TT on Conv layers is introduced in \cite{garipov2016ultimate}, where the 4D kernel tensor should be reshaped to size of $(H\cdot W)\times (C_1\cdot S_1) \times (C_2\cdot S_2) \times \cdots \times (C_d \cdot S_d)$ and the input feature maps are reshaped to $\mathbf{\mathcal{X}}\in \mathbb{R}^{H\times W\times C_1\times \cdots\times C_d}$. 
In the feedforward phase, the tensorized input $\mathbf{\mathcal{X}}$ will be contracted with each TT-core one by one. Although TT can significantly save storage costs, the computational complexity may be higher than the original Conv layer. Hence, HODEC (High-Order Decomposed Convolution) was proposed in \cite{yin2022hodec} to enable simultaneous reductions in computational and storage costs, which further decomposes each TT-cores into two 3rd-order tensors.

\subsubsection{Tensor Ring Decomposition}
\begin{figure}
    \centering
    \includegraphics[scale=0.4]{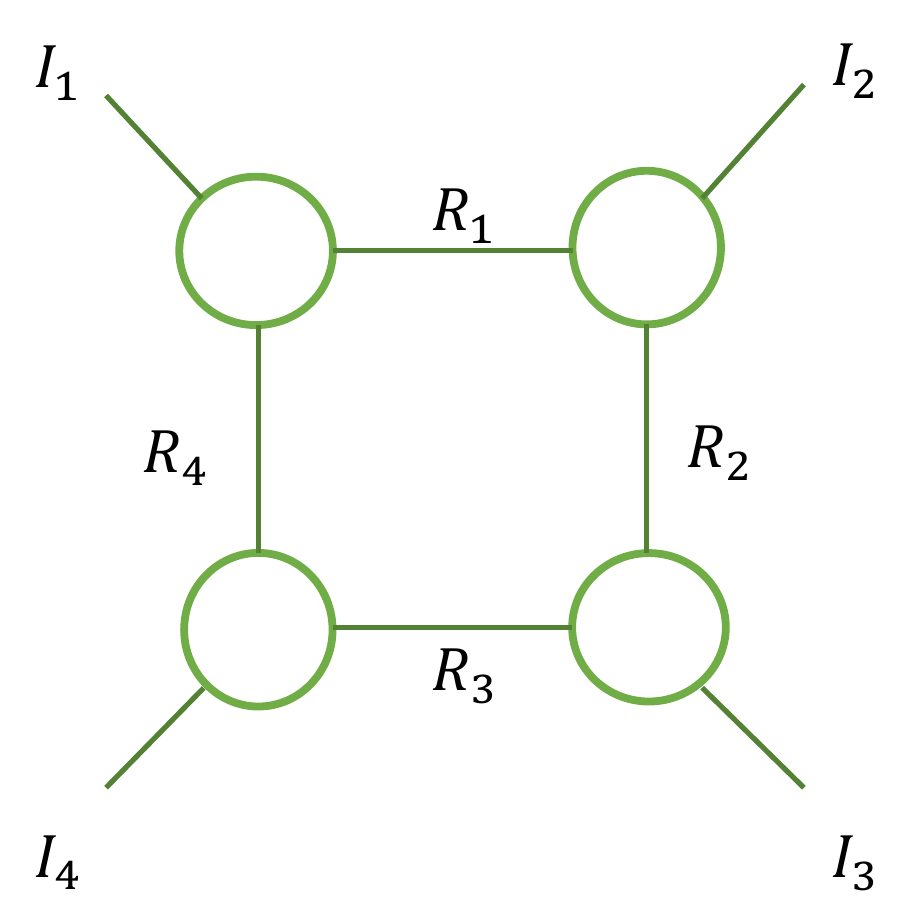}
    \caption{A 4th order tensor in TR format.}
    \label{TR}
\end{figure}

Due to the disunity of edge TT-cores, there is still an open issue that how to arrange dimensions of tensors to find the optimal TT format. To conquer this problem, Tensor Ring (TR) decomposition was proposed to perform a circular multilinear product over cores \cite{zhao2016tensor, huang2020robust, liu2020smooth, long2021bayesian}. Consider a given tensor, $\mathbf{\mathcal{X}}\in \mathbb{R}^{I_1\times I_2\times \cdots \times I_N}$, elementwise, we can formulate its  TR representation as:
\begin{equation}
\begin{split}
    \mathbf{\mathcal{X}}_{i_1, i_2, \cdots, i_N}= &\sum_{r_1,r_2,\cdots,r_N} \mathbf{\mathcal{G}}^1_{r_1, i_1, r_2}\mathbf{\mathcal{G}}^2_{r_2,i_2, r_3}\cdots\mathbf{\mathcal{G}}^N_{r_{N},i_{N}, r_1}\\
    = & tr(\sum_{r_2,\cdots,r_N} \mathbf{\mathcal{G}}^1_{:, i_1, r_2}\mathbf{\mathcal{G}}^2_{r_2,i_2, r_3}\cdots\mathbf{\mathcal{G}}^N_{r_{N},i_{N}, :}), \\
\end{split}
\end{equation}
where all cores $\{\mathbf{\mathcal{G}}^n \in \mathbb{R}^{R_{n}\times I_{n} \times R_{n+1}}\}_{n=1}^N$ with $R_{N+1}=R_1$ are called TR-cores. Its tensor diagram for a 4th order tensor is illustrated in Fig. \ref{TR}.
This form is equivalent to the sum of $R_1$ TT format. Thanks to the circular multilinear product gained by employing the trace operation, TR treats all the cores equivalently and becomes much more powerful and general than TT. 
\par Moreover, due to the circular strategy, TR amends the  variousness of  gradients in TT. Hence, TR is also suitable for compressing FC layers. In \cite{wang2018wide}, TR was first applied to compress DNNs. Specifically speaking, the weight matrix of FC layers should be reshaped into a 2$d$th order tensor of size $I_1\times \cdots\times I_d \times O_1\times\cdots\times O_2$, followed by representing the tensor into TR format. For the feedforward process, firstly, merge the first $d$ cores and the last $d$ cores to obtain $\mathbf{\mathcal{F}}_1\in \mathbb{R}^{R_1 \times I_1\times \cdots\times I_d \times R_{d+1}}$ and $\mathbf{\mathcal{F}}_2 \in \mathbb{R}^{R_{d+1}\times O_1\times \cdots\times O_d\times R_{1}}$, respectively. Then, we can calculate contraction between input $\mathbf{\mathcal{X}}\in \mathbb{R}^{I_1\times I_2\times \cdots\times I_d}$ and $\mathbf{\mathcal{F}}_1$, yielding a matrix that can  be contracted with $\mathbf{\mathcal{F}}_2$. The final output tensor will be of size $O_1\times O_2\times \cdots\times O_d$. For Conv layers, if  keeping the kernel tensor in 4th order and maintaining the spatial information, its TR-format can be formulated as:
\begin{equation}
    \mathbf{\mathcal{K}}_{s,c,h,w}=\sum_{r_1=1}^{R_1}\sum_{r_2=1}^{R_2}\sum_{r_3=1}^{R_3}\mathbf{\mathcal{U}}_{r_1, s, r_2}\mathbf{\mathcal{V}}_{r_2, c, r_3}\mathbf{\mathcal{Q}}_{r_3, h,w, r_1}.
\end{equation}
Hence, the original layer can be represented by three consecutive layers whose weight tensors are $\mathbf{\mathcal{V}}$, $\mathbf{\mathcal{Q}}$ and $\mathbf{\mathcal{U}}$ respectively. If a higher compression ratio is needed, we can further view $\mathbf{\mathcal{U}}$ and $\mathbf{\mathcal{V}}$ as tensors merged from $d$ core tensors respectively, with an extra computation burden of merging.

\subsubsection{Generalized Kronecker Product Decomposition}
Kronecker Product Decomposition (KPD) can factorize a matrix into two smaller factor matrices interconnected by Kronecker product, which has shown to be very effective when applied to compress RNNs \cite{thakker2019compressing}. To further compress Conv layers, it was generated to Generalized Kronecker Product Decomposition (GKPD) \cite{hameed2022convolutional}, which represents a tensor by the sum of multidimensional Kronecker product between two factor tensors. Formally, the multidimensional Kronecker product between $\mathbf{\mathcal{A}} \in \mathbb{R}^{J_1\times J_2 \times \cdots \times J_N}$ and $\mathbf{\mathcal{B}} \in \mathbb{R}^{K_1\times K_2 \times \cdots \times K_N}$ is formulated as:
\begin{equation}
    (\mathbf{\mathcal{A}}\otimes \mathbf{\mathcal{B}})_{i_1,i_2,\cdots,i_N}= \mathbf{\mathcal{A}}_{j_1,j_2,\cdots,j_N}\mathbf{\mathcal{B}}_{k_1,k_2,\cdots,k_N},
\end{equation}
where $j_n=\lfloor i_n/K_n\rfloor $ and $k_n= i_n$mod$K_n$. Based on this, for a given $N$th order tensor $\mathbf{\mathcal{X}} \in \mathbb{R}^{J_1K_1\times J_2K_2 \times \cdots \times J_NK_N}$, GKPD can be denoted as:
\begin{equation}
    \mathbf{\mathcal{X}}=\sum_{r=1}^{R}\mathbf{\mathcal{A}}_r\otimes \mathbf{\mathcal{B}}_r,
\end{equation}
where $R$ is referred to as Kronecker rank. For finding the best approximation in GKPD with a given $R$, we can transform this optimization problem into finding a best rank-$R$ approximation for a matrix, which can be solved by SVD conveniently, via carefully rearranging $\mathbf{\mathcal{X}}$ into a matrix and rearranging $\mathbf{\mathcal{A}}$ and $\mathbf{\mathcal{B}}$ into vectors.
\par For the realization of using GKPD to decompress Conv layers, the 4D kernel is represented as:
\begin{equation}
    \mathbf{\mathcal{W}}_{s,c,h,w}=\sum_{r=1}^{R}(\mathbf{\mathcal{A}}_r)_{s_1,c_1,h_1,w_1}\otimes (\mathbf{\mathcal{B}}_r)_{s_2,c_2,h_2,w_2},
\end{equation}
where $S_1 S_2=S$, $C_1 C_2=C$, $H_1 H_2=H$ and $W_1W_2=W$. The 2D convolution between each $\mathbf{\mathcal{A}}_r\otimes\mathbf{\mathcal{B}}_r$ and input can be transformed into a 3D convolution whose depth equals $C_2$, followed by multiple 2D convolutions. Furthermore, the group of $R$ Kronecker products can be viewed as $R$ parallel channels that calculate the above two steps separately. And it was analysed that large $S_1$ and $C_2$ can help to obtain more reduction in FLOPs.

\subsubsection{Semi-tensor Product-based Tensor Decomposition}
Semi-tensor product (STP) \cite{cheng2007survey} is a generation of the conventional matrix product, which extends the calculation of two dimensionally matching matrices to that of two dimensionally arbitrary matrices. Since tensor contraction is based on the conventional matrix product, we can further substitute STP into tensor contraction, which will lead to more general and flexible tensor decomposition methods. Proposed in \cite{zhao2021semi},  Semi-tensor Product-based Tensor Decomposition is designed to enhance the flexibility of Tucker decomposition, TT and TR by replacing the conventional matrix product in tensor contraction by STP, which demonstrates much higher efficiency than original methods. Consider a special case in which the number of columns  $\mathbf{X}\in\mathbb{R}^{M\times NP}$ is proportional to that of rows in $\mathbf{W}\in\mathbb{R}^{P\times Q}$, the STP can be denoted as:
\begin{equation}
    \mathbf{Y} = \mathbf{X}\ltimes \mathbf{W},
\end{equation}
or, elementwise, as:
\begin{equation}
    \mathbf{Y}_{m,g(n,q)} = \sum_{p=1}^{P}\mathbf{X}_{m,g(n,p)} \mathbf{W}_{p,q}.
\end{equation}
Note that $\mathbf{Y}\in\mathbb{R}^{M\times NQ} $, "$\ltimes$" denotes the STP,  $g(n, q)=(q-1)N+n$ and $g(n,p)=(p-1)N+n$ are re-indexing functions.
\par Hence, take STP-based Tucker decomposition as an example, namely STTu, which can be formulated as:
\begin{equation}
    \mathbf{\mathcal{X}}=\mathbf{\mathcal{G}}\ltimes_1 \mathbf{A}^{(1)}\ltimes_2 \mathbf{A}^{(2)}\ltimes_3\cdots\ltimes_N \mathbf{A}^{(N)},
 \end{equation}
where $\mathbf{\mathcal{G}}\in \mathbb{R}^{R_1\times R_2\times \cdots \times R_N}$, $\mathbf{A}^{(n)}\in \mathbb{R}^{\frac{I_n}{t}\times \frac{R_n}{t}}$. Compared with normal Tucker, the number of parameters is reduced from ($\prod \limits_{n=0}^N R_n + \sum_{n=1}^N I_nR_n$) to ($\prod \limits_{n=0}^N R_n + \sum_{n=1}^N \frac{I_nR_n}{t^2}$).
\subsection{Low Rank Optimization Method}
\par We have already introduced various tensor decomposition methods, but how to apply these methods to DNNs without significant accuracy degradation is an optimization problem, which remains to be discussed. Since the lower the tensor rank is, the higher compression ratio we will get, we hope that each layer of DNNs can be decomposed by very low rank tensor decomposition. However, as the rank gets lower, the approximation error increases, leading to a dramatic loss of accuracy. Hence, there is a tradeoff between accuracy and compression ratio, which is a widely studied problem. There are mainly three kinds of low rank optimization methods to achieve a good tradeoff: pre-train method, pre-set method and compression-aware method (representative works can be seen in Table \ref{three}). For each method, we will give the key points about the implementation in detail.

\begin{table*}[t]
\caption{Three types of low rank optimization method for model compression}
\label{three}
\begin{center}
\begin{tabular}{lcc}
\toprule
Method & Description & Representative Works \\
\midrule
Pre-train  & \tabincell{c}{pretrain the target model, apply tensor decomposition to\\ trained weight tensors, and then fine-tune to recover accuracy.} &  \cite{kim2015compression, liebenwein2021compressing, zhang2015accelerating, lebedev2014speeding} \\

Pre-set & \tabincell{c}{construct tensorized netwoks, set proper \\initialization, and then train the whole network.} & \cite{wang2018wide, garipov2016ultimate, ye2020block} \\

Compression-aware & \tabincell{c}{train the original network with normal optimizers \\ but enforce weight tensors to enjoy low rank structure.} & \cite{idelbayev2020low, yin2021towards, yin2022batude}\\
\bottomrule
\end{tabular}
\end{center}
\end{table*}

\subsubsection{Pre-train Method}
The pre-train method is the earliest used method in the literature of applying tensor compression to model compression, which directly decomposes an already trained network into a compact format, followed by fine-tuning to recover the accuracy. There are two critical issues for implementation: tensor rank selection and instability.
\par \textit{Tensor rank selection} means how to select the proper tensor rank of each layer in a network. Since the extent of redundancy varies from one layer to another, the rank of each layer is not supposed to be equal. Hence, unlike time-consuming trial-and-error, an efficient rank selection method should allocate the limited computation or storage resources to each layer reasonably via carefully deciding the rank of each layer, while ensuring the lowest accuracy degradation.
\par A simple but effective way is to set the rank of each layer to be proportional to the number of corresponding input or output channels, which usually performs better than roughly setting all ranks equal. A probabilistic matrix factorization tool called variational Bayesian matrix factorization (VBMF) \cite{nakajima2013global} is used in  \cite{kim2015compression} to estimate tensor ranks of a tensor in Tucker format. In order to get the mode-$n$ rank, the corresponding mode-$n$ matricization of the target tensor is viewed as an observation with noise. Then, VBMF is employed on the observation to filter out the noise and then obtain a low rank matrix. In \cite{zhang2015accelerating}, the rank selection problem is formulated as a combinatorial optimization problem \cite{reeves1993modern} with computation or memory resource constrained. The objective function is denoted as the product of PCA energy (the sum of singular values) of each layer, as the authors empirically observe that the PCA energy is roughly related to the classification accuracy. Similarly, the algorithm in  \cite{liebenwein2021compressing} also employs the idea that the approximation error is linked to the accuracy loss. But more efficiently and reasonably, it selects the maximum  approximation error of all the layers as a proxy for model accuracy. By minimizing this proxy, it is guaranteed that no layer decomposed will significantly reduce the accuracy. 
Together with the resource constraint, the final problem is a minimax optimization which can be solved by binary search.
\par Since the approximation error does not necessarily reflect the loss of accuracy, the above methods can only obtain a suboptimal rank configuration scheme. To address this challenge, reinforcement learning is employed to automatically select ranks \cite{cheng2020novel, samragh2019autorank}. In the established state-action-reward system, the reward favors a reduction in resource cost and penalizes loss of accuracy. The state (a possible global rank configuration of all the layers) that renders the maximum reward can be chosen as the next state. 
\par \textit{Instability} means that if a model is approximated by an unstable decomposed format such us CP format and TR format, it will lead to difficulty in fine-tuning, \ie{converge slowly and converge to a false local minima}. In \cite{mitchell1994slowly, harshman2004problem, krijnen2008non}, it was noted that there is a degeneracy problem that causes instability in CP decomposition. Specifically speaking, when CP represents a relatively high-rank tensor in a low-rank format, there are at least two rank-one components whose Frobenius norm goes to infinity and cancels each other out. Due to the instability, \cite{denton2014exploiting, lebedev2014speeding} fails to decompose the whole network by CP decomposition, as it is difficult to find a suitable fine-tuning learning rate. To deal with this challenge, \cite{astrid2017cp} proposes to use the Tensor Power Method \cite{allen2012sparse} to calculate CP decomposition and employ iterative fine-tuning, \ie{decompose one layer at a time and then fine-tune the entire network iterativelly}. The authors of \cite{phan2020stable} devise a procedure to minimize the sensitivity (a measure for the degeneracy degree) of the tensor reconstructed from CP format so that the decomposed network with low sensitivity can be fine-tuned faster and obtains a better accuracy. A more direct method proposed in \cite{veeramacheneni2022canonical} 
holds that each column of the factor matrix should be normalized after each update, as normalization can improve numerical stability in the outer vector product \cite{kolda2009tensor}. A similar instability problem also happens to TR \cite{espig2011optimization}. Hence, \cite{phan2022train} proposes a sensitivity correction procedure to address the problem via minimizing the sensitivity with an approximation error bounded constraint.

\subsubsection{Pre-set Method}
The pre-set method has the interpretation that a tensorized neural network that is preset to a low tensor rank format will be trained from scratch. As the method requires no pre-training, it can save a great deal of time to get a compressed model. However, the method is sensitive to initialization and difficult to achieve high accuracy due to the limited model capacity. Moreover, similar to the pre-train method, there are also problems in configuring ranks. In a nutshell, proper initialization and tensor rank selection are the main issues with this method. 
\par \textit{Initialization} plays an important role in providing a warm start for training DNNs \cite{glorot2010understanding} as well as for the training of low rank structure networks \cite{wang2018wide}, and can have an impact on the final accuracy to a large extent. An empirically determined suitable initialization distribution for weights in a layer is $\mathcal{N}(0,std=\sqrt{\frac{2}{N}})$, where $N$ denotes the total number of parameters in this layer. For a pre-set model, we should make sure that weights in each layer approximated by factor tensors also obey this distribution. For example, when a layer is compressed by TR and the distribution of each core tensor is $\mathcal{N}(0,\sigma^2)$, then
after merging these $d$ core tensors, elements of the merged tensor will have mean 0 and variance $R^d\sigma^{2d}$. Hence, we need to set $\sigma^2$ as $(\frac{2}{N})^{\frac{1}{d}}R^{-1}$ to obtain a good initialization. Similarly, for TT, the variance of TT-cores should be $(\frac{2}{N})^{\frac{1}{d}}R^{\frac{1}{d}-1}$. A more systematic analysis of initialization for any tensor decomposition method is introduced in \cite{pan2022unified}. It is suggested that by extracting the Backbone structure (\ie{a structure only contains contracted dimensions, since only the contraction operator will change the variance of weights}) from the original tensorized structure, an adjacency matrix can be obtained from node edges of the Backbone structure, which can be utilized to adjust the variance of factor tensors.
\par \textit{Tensor rank selection} is seldom studied in the works of training a tensorized neural network and usually set the ranks to equal in experiments, as it's difficult to verify the redundancy in each layer without a pre-training network. At present, there are only a few methods to solve this problem for specific tensor decompositions. Inspired by neural architecture search (NAS) \cite{zoph2016neural}, \cite{li2022heuristic} proposes a progressive searching tensor ring network (PSTRN), which has the ability to find an appropriate rank configuration for TR efficiently. In this algorithm, an evolutionary phase and a progressive phase are alternatively performed. While the evolutionary phase is responsible for deriving good rank choices within the search space via multi-objective genetic algorithm NSGA-II \cite{deb2002fast}, the progressive phase is responsible for narrowing the search space in the vicinity of the optimized rank coming from the previous evolutionary phase. For rank selection with TT decomposition, \cite{hawkins2021bayesian} proposes a low-rank Bayesian tensorized neural network. Bayesian methods are always used to infer tensor ranks in CP format or Tucker format through low-rank priors in tensor completion tasks\cite{rai2014scalable,guhaniyogi2017bayesian,bazerque2013rank}. This paper generates this approach to TT format and nonlinear neural networks.

A more easily implemented method, modified Beam-search, was proposed in \cite{eo2023effective} to find the optimal rank setting, costing much lower search time than the full search. To verify optimality, it adopts the validation accuracy on a mini-batch validation dataset as its metric. This method is applicable to all kinds of tensor decompositions.

\subsubsection{Compression-aware Method}
Compression-aware method is the method that through standard training and iterative optimization, the weights of kernels and FC layers can gradually have desired low tensor rank structures. That is, consider the future compression into the standard training phase. Upon the end of this one-shot training, the suitable tensor ranks are automatically learned, without efforts to design efficient rank selection schemes. Moreover, since the training process is still on the original network structure instead of a deeper factorized network, it's easy to converge towards high accuracy without being prone to gradient vanishing or explosion. There are mainly two kinds of ways to realize this idea, namely using low rank regularization and solving constrained optimization.
\par \textit{low rank regularization} is similar to the sparse regularization which is always used in DNNs to avoid overfitting. The main idea of low rank regularization is to add low rank regularizer on weights in each layer to the basic loss function. Hence, with the constraint of such regularizer, weight tensors will gradually have a desired low rank structure during training. Then, after low rank approximation, there is no need to retrain for a long time and no risk of unstable recovery.

\par For the low rank regularizer, an index to measure the low rank degree is essential. Since explicitly minimizing the rank of a matrix is NP-hard, nuclear norm \cite{cai2010singular} was widely used as a continuous convex relaxation form of rank. 
In \cite{alvarez2017compression}, the sum of nuclear norms of weight matrices in each layer is added to cross-entropy loss, yielding a new optimization problem which can be solved by proximal stochastic gradient descent. Similarly, \cite{xu2019trained} also uses nuclear norm and the same optimization problem is solved by stochastic sub-gradient descent \cite{avron2012efficient}. In addition, this paper embeds the low rank approximation into the training phase to boost the low rank structure. 
\par However, for the above, SVD will be performed on every training step, which is inefficient, especially for larger models. Hence, \cite{yang2020learning} proposes SVD training which performs training directly on the decomposed factors. By employing sparsity regularization on singular values, it can achieve the goal of boosting low rank. In order to maintain the valid SVD form, orthogonality regularization on the left and right singular matrices is necessary. Moreover, Orthogonality also can efficiently prevent the gradient to explode or vanish, therefore achieving higher accuracy.
\par \textit{Solving constrained optimization} is a method that through solving an optimization problem with explicit or implicit constraints on tensor ranks of weights, we can get an optimal network not only with low loss but also with low rank structures. Classically, \cite{idelbayev2020low} forms the low rank constrained problem as minimizing the sum of the loss and a memory/computation cost function but constraining each rank not to exceed a maximum rank.
It can be solved by a learning-compression algorithm \cite{carreira2018learning}. More conveniently, \cite{yin2022batude} directly uses budget (\eg{memory/computation cost}) as constraints, with low rank regularizer added on the loss function. However, since it represents tensor ranks by the sum of nuclear norms of unfolding matrices in each mode, it cannot be generalized to certain decomposition methods such as CP and BTD. And when dealing with high-order tensors, there will be too many auxiliary variables used in the augmented Lagrangian algorithm, which will affect convergence. Without using nuclear norm, \cite{yin2021towards} just set the upper bound of ranks, therefore it is applicable to various tensor decompositions. 

\par The above methods have an unsatisfactory tradeoff between accuracy and compression. To address this drawback, the Frank Wolfe algorithm is utilized in \cite{zimmer2022compression} to optimize network weights with the low-rank constraint. This improvement benefits from the highly structured update directions of Frank Wolfe.

\par For compression-aware methods, using different sparsity measures as low rank regularizers will greatly impact compression performance. For an instance, it was noted in \cite{yang2020learning} that the $\ell^1$ measure (\eg{nuclear norm}) is more suitable for an extremely high compression ratio while Hoyer measure performs better when aiming for a relatively low compression ratio. Hence, it's essential to dig out an efficient sparsity measure that is attractive for any compression ratio. This is exactly the point we want to make below.  

\subsection{Sparsity Measure}
\par Recently, researches on compression-aware method emerge in large numbers and plenty of experiments show that with the premise of using the same tensor decomposition method, compression-aware method can outperform the other two methods\cite{yin2022batude, yin2021towards,yang2020learning}. Hence, we should pay more attention to it. One thing that has not been fully studied is the sparsity measure used. As the most classical convex relaxation form of rank, nuclear norm ($\ell_1$ measure) is widely used. However, there is no evidence that the nuclear norm is a perfect choice. Consequently, a comparison between common sparsity measures should be made. Finding a more efficient measure may greatly improve the compression capability of existing compression-aware algorithms.
\subsubsection{Common sparsity measure}
\par For sparse representation problems, the $\ell_0$ norm defined as the number of non-zeros is the traditional measure of sparseness. However, Since the $\ell_0$ norm is sensitive to noise and its derivative contains no information, the $\ell_p$ norm with $0< p\leq 1$ is introduced to less consider the small elements \cite{shi2019sparse}. For a vector $\mathbf{x}\in \mathbb{R}^{N}$, its $\ell_p$ norm can be formulated as:
\begin{equation}
    \ell_{p}(\mathbf{x})= (\sum_{i=1}^{N}|x_{i}|^p)^{\frac{1}{p}}.
\end{equation}

The $\ell_1$ norm,  $\ell_p$ norm with $p = 1$, is the most widely used sparsity measure. Formally, consider a vector $\mathbf{x}\in \mathbb{R}^{N}$, its $\ell_1$ norm can be denoted as:
\begin{equation}
    \ell_{1}(\mathbf{x})= \sum_{i=1}^{N}|x_{i}|.
\end{equation}
The $\ell_1$ norm is introduced in \cite{4100845} as a more practical substitute for the $\ell_0$ norm. In addition, in order to better measure sparsity in noisy data, more flexible forms based on $\ell_1$ norm are proposed in \cite{bogdan2013statistical,huang2015two}, namely sorted $\ell_1$ norm and two-level $\ell_1$ norm. The sorted $\ell_1$ norm is formulated as:
\begin{equation}
    \ell_{1}^{\text{sort}}(\mathbf{x})= \sum_{i=1}^{N}\lambda_i |x_{i}|,
\end{equation}
where $\lambda_1\geq\lambda_2\geq \cdots\geq\lambda_N\geq 0$. In this way, the higher the magnitude of the element, the larger the penalty on it. More concisely, the two-level $\ell_1$ norm only considers two levels of penalty, which can be formulated as:
\begin{equation}
    \ell_{1}^{\text{2level}}(\mathbf{x})= \rho\sum_{i\in \mathbb{I}_1} |x_{i}|+\sum_{j\in \mathbb{I}_2} |x_{j}|.
\end{equation}
Here, we have $|x_i|\geq |x_j|, \forall i\in \mathbb{I}_1, \forall j \in \mathbb{I}_2$.
\par
\begin{figure}
    \centering
    \includegraphics[scale=0.45]{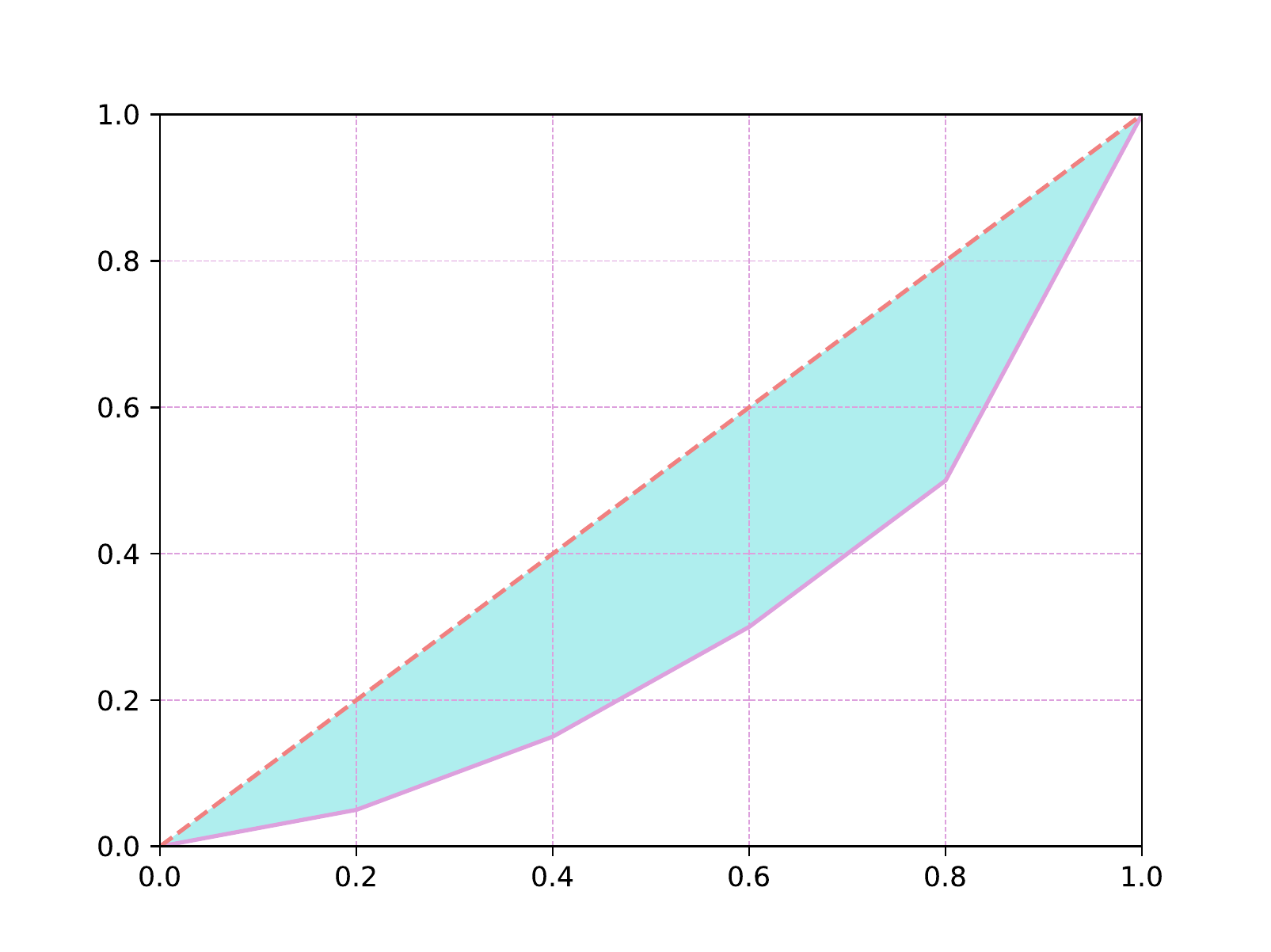}
    \caption{A graphical illustration of Gini Index for a vector [1, 2, 3, 4, 10]. The dot line (45-degree line) represents the case in which all elements are equal, and the full line is the Lorenz curve of the vector. Twice the area between them is equal to the Gini Index of such a vector.}
    \label{gini}
\end{figure}

 The Gini Index was initially proposed as a measure of the inequity of wealth \cite{dalton1920measurement,lorenz1905methods}. Afterward, the utility of Gini index as a measure of sparsity has been demonstrated in \cite{rickard2006sparse,hurley2005parameterized}. Given a sorted vector $\mathbf{x}\in \mathbb{R}^{N}$ whose elements increases by degrees, \ie{$x_1\leq x_2 \leq \cdots \leq x_N$}, its Gini Index is given by:
\begin{equation}
    G(\mathbf{x})=1-2\sum_{i=1}^{N}\frac{x_i}{||\mathbf{x}||_1}(\frac{N-i+\frac{1}{2}}{N}).
\end{equation}
Note that if all elements are equal, \ie{no sparsity}, the Gini Index reaches its minimal 0. For the most sparse case, \ie{only $x_N$ is non-zero}, the Gini Index goes to a maximum of $1-\frac{1}{N}$. Graphically, the Gini Index can be represented as twice the area between the Lorenz curve \cite{lorenz1905methods} and the 45-degree line. Each point on the Lorenz curve ($x=a$, $y=b$) has the interpretation that 
top $100\times a$ percent of the sorted elements expresses $100\times b$ percent of the total power. The degree line represents the least sparse case with Gini Index 
equal to 0. Fig. \ref{gini} illustrates the Lorenz curve for a vector.  

\par The Hoyer measure was devised in \cite{hoyer2004non} as a new sparsity measure, which is a normalized version of $\frac{\ell_2}{\ell_1}$. For a given vector $\mathbf{x}\in \mathbb{R}^{N}$, its Hoyer measure can be formulated as:
\begin{equation}
    H(\mathbf{x})=\frac{\sqrt{N}-\frac{||\mathbf{x}||_1}{||\mathbf{x}||_2}}{\sqrt{N}-1}.
\end{equation}
This function goes to unity if and only if $\mathbf{x}$ contains only a
single non-zero component, and takes a value of zero if and only if all components are equal, changing smoothly between the two extremes.

\par The above-mentioned sparsity measure can be applied to the singular value vector as a low rank measure of the corresponding matrix. There are other non-strict measures for the rank of a matrix. Here, we concentrate on effective rank \cite{roy2007effective}. Let us consider a matrix $X$ of size $M\times N$ whose singular value vector is denoted by $\mathbf{\sigma} \in \mathbb{R}^{K}$ with $K=\min \{M,N\}$, its effective rank can be given by:
\begin{equation}
    E(\mathbf{X}) = \operatorname{exp}(-\sum_{i=1}^{K}\overline{\sigma}_{i}\operatorname{log}\overline{\sigma}_{i}), 
\end{equation}
where $\overline{\sigma}_{i}=\frac{\sigma_{i}}{||\mathbf{\sigma}||_1}$, the logarithm is to the base $e$, and the convention that $\operatorname{log}0 = 0$ is adopted. This measure is maximized when all the singular values are equal, and minimized when the maximum singular value is much larger than other values.

\subsubsection{Comparison}
In the compression-aware method, it is common to employ sparsity regularizer on singular value vectors to encourage weight matrices to lie in a low rank subspace. The nuclear norm is the most frequently used. However, it simply makes everything closer to zero, which is unfriendly to keeping the energy of weight matrices. Hence, we prefer other measures that encourage the insignificant singular values (with small magnitude) to go to zero but keep the significant values (with large magnitude) or make them larger to maintain the energy. Hence, we choose Gini Index, Hoyer, and effective rank as potential objects, and make a comparison among them together with the nuclear norm. 
\par We execute the comparison experiment on ResNet32 trained on the Cifar10 dataset. We utilize the most simple SVD to compress the network, and in the compression-aware training phase, we employ various sparsity measures on singular vector values of each weight matrix,  with a hyperparameter $\lambda$ to make the balance between accuracy and low rank. After this training, there are many singular values close to zero that can be set to zero without degrading performance. An appropriate indicator for identifying singular values retained is introduced in \cite{chen2020deep}, namely spectral norm based indicator. This indicator is defined as 
the ratio of the largest discarded singular value to the maximal singular value. It is more efficient than the normal Frobenius norm based indicator \cite{osawa2017evaluating}, as it can get rid of the interference caused by small and noisy singular values. 
\begin{figure}[!htbp]
    \centering
    \includegraphics[scale=0.6]{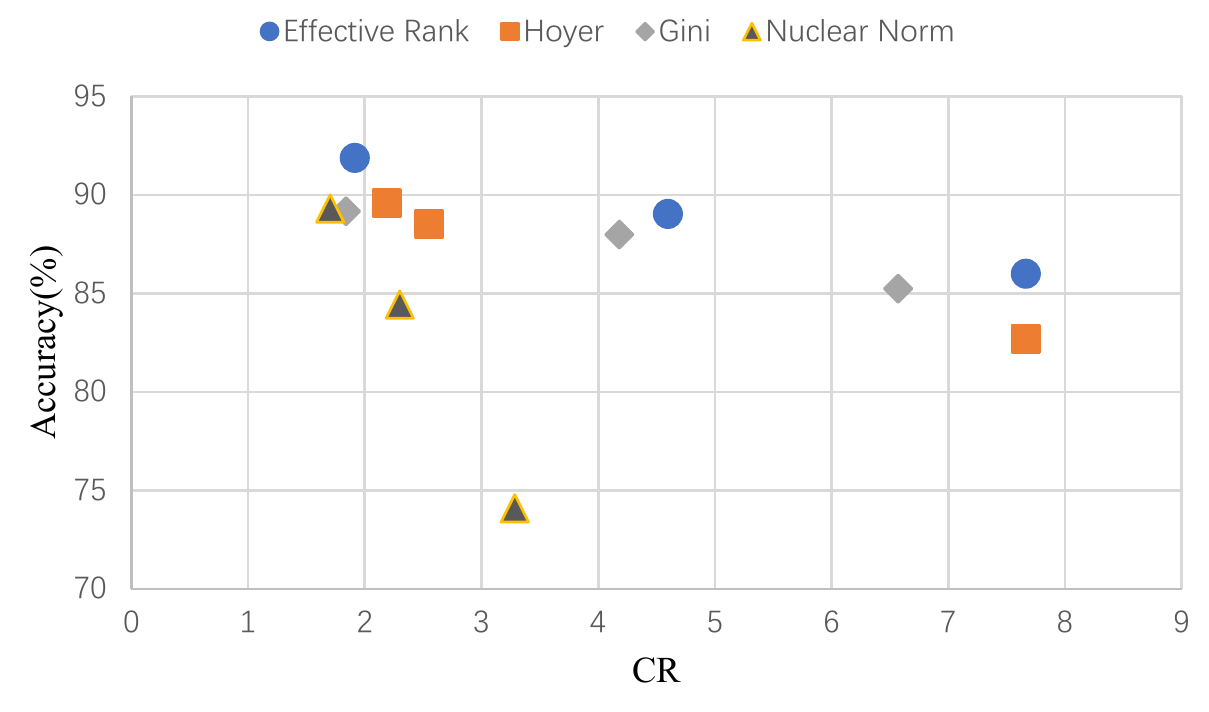}
    \caption{Accuracy on Cifar10 vs compression ratio of the number of parameters in ResNet32. Effect of different sparsity measures.}
    \label{compare}
\end{figure}
\par Fig. \ref{compare} shows the effect of the four sparsity measures. The most frequently used Nuclear Norm shows the worst performance. With the increase in the compression rate, the accuracy drops sharply. The reason behind this can be that at the time of pursuing a high compression ratio, the value of $\lambda$ is increased, with more singular values imposed to zero. It dramatically destroys the expressive ability of the model. This figure also suggests that effective rank surpasses the rest measures for any compression regime. To be specific, when the accuracy is close to 85\%, effective rank can achieve a compression ratio almost 4$\times$ greater than the nuclear norm. And in the case of 90\%, it can achieve  2$\times$ greater than Hoyer. For a low compression regime, effective rank also has the greatest potential to achieve accuracy close to the original.

\par For the spectral norm based indicator, if we are in the need of discarding most of singular values to gain a high compression ratio, there are two choices: increase the value of maximum singular value or decrease the value of tiny singular values. However, increasing the value of the maximum singular value 10 times is much more difficult than decreasing the value of tiny singular values 10 times. Hence, we prefer a measure that can strongly encourage tiny singular values to reach 0. This is also the reason why effective rank can demonstrate great efficiency. 

\section{Integratable Techniques} \label{sec:3}
\par Apart from low rank approximation, there are other compression schemes that can result in a significant reduction of parameters at the expense of only a small drop in output accuracies, such as pruning  \cite{blalock2020state}, weight-sharing \cite{chen2015compressing},
sparsification \cite{han2015learning} and knowledge distillation \cite{han2015deep}. Undoubtedly, the integration of these parameter reduction techniques, namely parallel integration, can further enhance the efficiency of compression.
While plenty of surveys suggest integrating various compression techniques, a detailed discussion on the combination between low rank approximation and other schemes is still lacking. In addition, not only the reduction of parameters but also the reduction of bits for representing parameters can significantly cut down the high complexity, which can be realized by quantization and entropy coding. Quantization can represent each parameter with lower bit-width, and entropy coding can use codewords to encode source symbols. Both techniques are orthogonal to the above parameter reduction methods. Hence, we can directly employ them on 
a compact model to gain a more compact representation, namely orthogonal integration. 
Table \ref{integration} lists representative works of different types of integration, and Table \ref{C_A} lists whether these techniques can compress or accelerate models.
\begin{table*}[htbp]
\caption{Integratable Techniques}
\label{integration}
\begin{center}
\resizebox{\textwidth}{!}{
\begin{tabular}{l|ccc}
\hline
 &Techniques & Description & Representative Integration Works \\
\hline
\multirow{4}{*}{Parallel Integration} & pruning & \tabincell{c}{discard insignificant connections} & \cite{chen2020deep, alvarez2017compression, ruan2020edp} \\
& sparsification & \tabincell{c}{zero out insignificant weights} &  \cite{swaminathan2020sparse, liu2015sparse, wen2020learning}\\
& weight sharing & \tabincell{c}{share weights across different connections} & \cite{obukhov2020t, li2019learning, sun2020deep} \\
& knowledge distillation & \tabincell{c}{transfer knowledge learned from teacher to student} & \cite{li2020few, lin2018holistic, 2020Knowledge} \\
\hline
\multirow{2}{*}{Orthogonal Integration} & quantization & \tabincell{c}{reduce precision} & \cite{lee2021qttnet, kuzmin2022quantized, nekooei2022compression} \\
& entropy coding & \tabincell{c}{encode weights into binary codewords} &  \cite{choi2020universal, wiedemann2020deepcabac, chen2021efficient} \\
\hline
\end{tabular}}
\end{center}
\end{table*}

\begin{table}[!ht]
    \caption{Ability to compress and accelerate for various techniques}
    \label{C_A}
    \centering
    \begin{tabular}{c|l|l}
    \hline
     Techniques  & Acceleration &  Compression\\
     \hline
     pruning & \Checkmark & \Checkmark \\
     sparsification & \Checkmark & \Checkmark \\
     weight sharing & \XSolid & \Checkmark \\
     knowledge distillation & \Checkmark & \Checkmark \\
     quantization & \Checkmark & \Checkmark \\
     entropy coding & \XSolid & \Checkmark \\
     \hline
    \end{tabular}

\end{table}

\subsection{Parallel Integration}
\par In this part, we will give an all-round survey on how to integrate low rank approximation with other parallel compression techniques, including pruning, weight sharing, sparsification, and knowledge distillation. Through joint-way use, we can pursue a higher compression capacity.
\subsubsection{Integration with Pruning}
Pruning is used to find unimportant connections in a full structure network and then abandon them, resulting in a compact structure without significant loss of accuracy. Pruning can be classified according to various levels of granularity, including weight-level, filter-level and layer-level. Weight-level is the most flexible approach \cite{han2015learning} and can gain the lowest memory costs by storing in sparse matrix format such as Compressed Sparse Column (CSC) \cite{han2015deep}. However, it leads to difficulty in inference due to the need for identifying each weight kept or abandoned. That is, this approach cannot speed up inference or save the memory footprint unless supported by hardware \cite{han2016eie}. Layer-level aims at abandoning trivial layers, which is unsuitable for shallow networks \cite{chen2018shallowing}. To overcome these drawbacks, a more flexible and applicable approach, namely filter-level, is proposed \cite{huang2018learning}. Filter-level considers each filter as a unit and discards insignificant filters to obtain a compact model but with regular structures. Note that for two successive Conv layers, the removal of a filter in the first kernel leads to the removal of the input channel in the next kernel.
\par Filter pruning doesn't deal with the redundancy within a filter, while low rank approximation can overcome this by representing each filter in low rank format. Hence, it is promising to combine them to explore a higher compression ratio. \cite{goyal2019compression} proposes to perform filter pruning first and then employ Tucker decomposition on the pruned kernels. Experiments in \cite{goyal2019compression} show that the joint-way approach can achieve up to 57\% higher compression ratio than either of them. \cite{chen2020deep} exchanges the order of filter pruning and low-rank approximation since the smaller filters obtained by low rank approximation can reduce the probability of discarding essential filters. In addition, previous works pointed out that filter pruning is likely to prune more filters in deeper layers, resulting in still high computation costs of the whole network\cite{molchanov2016pruning}. But with the help of low rank approximation, the shallow layers also can be compressed. Then, both high-level compression of memory and computation costs can be achieved.
\par One branch of works can achieve low rank approximation and filter pruning simultaneously via regularizers. In \cite{alvarez2017compression}, the nuclear norm regularizer and the sparse group Lasso regularizer \cite{alvarez2016learning} are combined to make weight matrices not only low rank but also group sparse. Then the original layer can be represented by two smaller layers, followed by discarding insignificant input channels of the first layer and output channels of the second layer. Different from this method, \cite{ruan2020edp} uses one type of regularizer to achieve both two motivations. It represents a weight matrix by a basis matrix and a coefficient matrix. By imposing $\ell_{2,1}$ regularization both on the coefficient matrix and its transpose, the basis matrix can turn to be low rank and insignificant output channels are identified. Or, there are also some works that employ the two techniques on different modules of a network. For instance, aiming for Transformer architecture,  \cite{kumar2022vision} compresses the attention blocks by low rank approximation and applies to prune to feedforward blocks, which gains great enhancement.

\subsubsection{Integration with Sparsification}
\par Sparsification in DNNs focuses on making  weight matrices sparser so that sparse matrix computation can be employed to reduce high computation costs. Meanwhile, it can provide storage efficiency, as non-zeros and their locations can be recorded in Compressed Sparse Row (CSR) \cite{han2015deep} or Ellpack Sparse Block (ESB) \cite{liu2013efficient} format. There are two types of sparsification, namely irregular sparsity and structural sparsity. When the non-zeros are located randomly in the matrix, we call it irregular sparsity, which is flexible but may result in poor acceleration due to its irregular data access pattern. On the contrary, structural sparsity can achieve regular data access patterns. To be more specific, structural sparsity normally zeros out a series of continuous elements in the matrix. 
\par Low rank approximation factors a matrix into smaller components, but these components still contain tiny elements which can be zeroed out without leading to a significant increase in approximation error. Hence, it's promising to combine low rank approximation and sparsification to achieve better compression. Sparse Principal Component Analysis (SPCA) \cite{zou2006sparse} is a well-known instance to integrate factorization with sparsity. The main idea of SPCA is to make each principal component only contain a few features of data, so that SPCA is more explainable than PCA. There are also sparse HOSVD and sparse CP proposed in \cite{allen2012sparse}.

\par In \cite{liu2015sparse}, it has shown that surprisingly high sparsity can be achieved after two-stage decomposition. It was claimed that more than 90\% of parameters can be zeroed out with less than 1\% accuracy degradation on ILSVRC2012 dataset. In this algorithm, sparsity and low rank are achieved by employing $\ell_1$ norm and $\ell_{2,1}$ norm respectively on a coefficient matrix. Finally, it converts the convolution operation in Conv layers into spare matrix multiplication, which dramatically reduces computation costs. Sparse SVD, \ie{factor matrices in SVD are sparsed}, was proposed in \cite{swaminathan2020sparse}, which outperforms truncated SVD. According to the view that a portion of the input and output neurons in a layer may be insignificant, the corresponding rows of the left and right singular matrix can be zeroed out. And considering the importance of entries in a row of left or right singular matrix decreases from left to right, the Sparse SVD prefers to abandon entries nearing the right. The resulting structural sparsity allows BLAS\cite{lebedev2016fast} libraries to be used for higher speed-up.
\par Aiming for RNNs, \cite{wen2020learning} proposed Low-Rank Structured Sparsity. Considering dimensional invariance in time, this method employs $\ell_1$ regularization on the left and right singular matrix derived from SVD, resulting in a column-wise and row-wise sparse matrix without dimension distortion. 
\subsubsection{Integration with Weight Sharing}
Weight sharing is defined as an operation that shares parameters across different connections in DNNs by exploiting redundancy. In order to design a more complex network with a better capacity for feature extraction, it is common to copy or reform some well-designed modules in a shallow network, and then add new modules to the end, yielding a deeper network. One typical network is the well-known ResNet \cite{lecun2015deep}. 
Due to this similarity, it is promising to explore a more compact representation by sharing parameters across these similar subnetworks. For low rank approximation, similarly, the idea of sharing factor tensors across tensor decompositions of similar weight tensors can also be adopted.
\par A simple illustration of integration with weight sharing can be found in \cite{li2019learning}, where a set of 3D filter bases is shared across several or all convolutional layers. The search for bases is equivalent to low rank approximation of all the matrix-shaped kernels with a shared basis matrix.
\par Some tensor decomposition methods naturally combine weight sharing. For example, in the previously mentioned Semi-tensor Product-based Tensor Decomposition, STP can calculate a multiplication between a vector $\mathbf{x}\in \mathbb{R}^{N}$ and a weight vector $\mathbf{w}\in\mathbb{R}^{P}$, resulting an output vector $\mathbf{y}\in\mathbb{R}^{\frac{N}{P}}$. The $\frac{N}{P}$ entries in each block of $\mathbf{x}$ share one weight parameter of $\mathbf{W}$. 
\par Alternatively, one branch of works shares factor tensors across tensor decompositions of weight tensors in different layers. \cite{obukhov2020t} proposed T-Basis, which constructs a set of 3rd-order tensors. For an arbitrary-shaped tensor, each of its TR-cores can be represented as a linear combination of T-Basis. Hence, a compact representation of DNNs can be derived. \cite{sun2020deep} proposed coupled TT, which contains a common component and an independent component. The common component is represented by shared TT-cores for similar network blocks, while the independent components in TT format are various from different layers to maintain the characteristics of each layer. 

\subsubsection{Integration with Knowledge Distillation}
Knowledge distillation \cite{hinton2015distilling} is a promising solution, which aims to feed some extra knowledge learned from teacher networks (one or more complex networks) into a student network (much simpler network). With the help of a teacher, the student can achieve comparable accuracy but with much lower memory and computation costs compared with the teacher. Let $\mathbf{q_s}$ and $\mathbf{q_t}$ denote the softmax outputs of the student network and teacher network, respectively. The student network will be trained via aligning $\mathbf{q_s}$ and $\mathbf{q_t}$. But in the case that $\mathbf{q_t}$ is close to the one-hot code of true labels, the information contained in small values cannot be transferred to the student. Hence, a trick named temperature \cite{hinton2015distilling} is utilized to soften the distribution of both $\mathbf{q_s}$ and $\mathbf{q_t}$.
\par Networks compressed by low rank approximation is also a simpler network that can learn knowledge from the uncompressed version. In general, the decomposed networks are recovered by simply fine-tuning to minimize the cross-entropy function. However, the fine-tuning process always converges slowly and cannot recover the original accuracy well. Hence, this underlines the need for training the compressed network with information from the corresponding pre-training network.
\par However, it was demonstrated in \cite{glorot2010understanding} that it is difficult to train a student network deeper than the teacher network with knowledge distillation due to the undesirable phenomenon of vanishing gradient. Hence, a novel knowledge transfer (KT) was proposed in \cite{lin2018holistic}, which aligns both outputs and intermediate responses from a teacher (original) network to its student (compressed) network. Experiments show that it surpasses the common fine-tuning and knowledge distillation,  particularly with a high compression ratio.
\par However, the KT method is still time-consuming and has a demand for a fully annotated large-scale training set, which may be infeasible in practice. Li \etal \cite{li2020few} proposed a revised knowledge distillation that only requires a few label-free samples. It adds a 1×1 Conv layer at the end of each block of the student network, and aligns block-level outputs of teacher and student by estimating the 1×1 Conv layer’s parameters using least-squared regression. Since the number of parameters in 1×1 Conv layers is relatively small, only a few samples are necessary. It also enables fast model convergence, thereby saving much time for recovery of accuracy. After learning, the 1×1 Conv layer will be merged into the previous layer, without an increase in the number of parameters. 

\subsection{Orthogonal Integration}
\subsubsection{Quantization}
The operation that maps data from full precision to reduced precision is referred to as quantization. In the training and inference phase of DNNs, it is common to represent weights and activations in 32-bit. However, transferring data in 32-bit is a burden, and Multiply-Accumulate (MAC) will be operated between 32-bit floating-point values. In addition, energy consumed scales linearly to quadratically with the number of bits used. Hence, lowering the precision is necessary for the reduction of memory size, acceleration and energy saving.
\par There are some special advantages of applying quantization on neural networks. First, compared with continuous form, the discrete representations are more robust to noise \cite{chaudhuri2016computational,faisal2008noise} and are more similar to the way of storing information in human brains \cite{vanrullen2003perception, tee2020information}. Second, both high generalization power  \cite{khaw2017discrete,latimer2015single} and high efficiency under limited resources \cite{varshney2006optimal} of discrete forms are actually what deep learning needs. Third, common compression methods, like low rank approximation, weight-sharing, and pruning, focus on either memory compression or acceleration so that it is deficient to achieve significant acceleration and compression simultaneously for a whole network, while quantization can conquer this challenge. In addition, it is shown in \cite{lin2016fixed} that most of the weights and activations in DNNs are close to zero, which can greatly promote the compression ability of quantization. A more detailed survey about implementing quantization on DNNs can be found in \cite{gholami2021survey, nagel2021white}. 
\par A straight-forward way to combine low rank approximation and quantization is to consider the network compressed by tensor decomposition as a new network, which can be normally further compressed by various quantization methods. However, since there is already an approximation error derived from decomposition, the subsequent quantization will suffer from serious accuracy degradation. Hence, a novel integration method that considers low rank decomposition and quantization simultaneously instead of successively has the potential to address the challenge. 
\par This idea can be found in \cite{kozyrskiy2020cnn}, where both factors of Tucker format and activations are quantized, and with the help of knowledge distillation, the approximation error is minimized. In \cite{kuzmin2022quantized}, quantization is introduced in principal component analysis (PCA), where the component matrix and the coefficient matrix are quantized with different bit-widths. Together With a sparsity constraint on the coefficient matrix, the approximation error on the data manifold derived from low rank decomposition, sparsity and quantization will be minimized by an iterative projected gradient descent method. 
\par Also, there are some approaches that directly extend basic tensor decomposition algorithms to tensor decompositions with quantized 
factors. For instance, quantized CP-ALS was proposed in \cite{nekooei2022compression}, wherein each optimization iteration factors are quantized, and it is shown that the reconstruction error under ALS and quantized ALS are almost the same.
\par The above-mentioned methods are all aiming at approximating a tensor with quantized factors, which is not suitable for pre-set method. In \cite{lee2021qttnet}, a quantized tensor train (QTT) is utilized for compressing three-dimensional convolutional neural networks. TT-Cores in tensorizied neural networks are first quantized, and then the quantization of feedforward process is also made, achieving a 3× faster inference than using only TT.

\subsubsection{Entropy Coding}
Entropy coding is a lossless compression scheme, which encodes source symbols with a lower number of bits per symbol by exploiting the probability distribution of source
 \cite{recanatesi2019dimensionality}. Entropy coding originally adopted for data compression is introduced to further reduce the memory size of quantized DNNs by representing quantized weights with binary codewords \cite{han2015deep}. It uses Huffman coding to further save 20\% to 30\% of network storage with no loss of accuracy. 
 \par Huffman coding is a theoretically optimal method to encode multivariate independence source symbols, but with the precondition that statistical characteristics of source symbols are already known. 
 There is a problem with DNNs that statistical characteristics of weights calculated by histogram is a time-consuming preparation and are different for each network, even for a network fine-tuned. Hence, an encoding method without the need for exact statistics is more efficient for compressing DNNs. 
\par One branch of works called universal coding, such as the variants of Lempel–ZivWelch \cite{ziv1977universal,ziv1978compression,welch1984technique} and the Burrows–Wheeler transform \cite{effros2002universal}, can be applied to deal with this problem. The 'universal' means that this coding method has a general probability model which can be slightly adapted to a broad class of input sources. In application, Deep Context-based Adaptive Binary Arithmetic Coder (DeepCABAC) \cite{wiedemann2020deepcabac}, one type of universal coding, is utilized to encode weights in DNNs. It is the first attempt to apply state-of-art video coding methods (CABAC) to DNNs. Compared with Huffman coding, DeepCABAC also has the advantage of higher efficiency in throughput. 
\par However, both Huffman coding and DeepCABAC are Fixed-to-Variable (F2V) schemes in which the number of bits for each symbol is variable. Due to the variable length in codewords, it is inefficient for memory usage when decoding, and hence leads to high latency for inference. Instead, Tunstall coding \cite{chen2021efficient}, a Variable-to-Fixed (V2F) method, is designed to fix the length of each codeword so that we can process multiple bits simultaneously and decode multiple encoded strings in parallel. It is reported that Tunstall coding can achieve around 6$\times$ faster decoding than Huffman coding.

\section{Low Rank Optimization for Subspace Training} \label{sec:4}

\subsection{Low Rank Function}
\par For a differentiable real-valued function, if its gradient always lies in a fixed low-dimensional subspace, it can be called a low rank function \cite{cosson2022gradient}. The dimensionality of such subspaces is much lower than the number of independent variables, and it is referred to as the rank of the function. Ridge functions are the most common low rank function, which are defined as functions that can be converted into a univariate function by applying an affine transformation to the argument \cite{logan1975optimal}. Hence, the gradient of such a function can also be projected into a line. For example, the least-square regression function which is a classic ridge function can be considered a rank-one function. The low rank property of ridge functions makes them widely used in classic statistics. They are utilized as regression functions in projection pursuit regression to deal with the curse of dimensionality and the noise in data \cite{donoho1989projection}. In scientific computing, since the variables of functions for uncertainty quantification are always correlated, the concept of active subspaces can be utilized to 
reveal a set of independent variables whose fluctuation can lead to the most significant change  \cite{constantine2015exploiting, liu2013multi}.
\par Low rank property has also been found in the training phase of DNNs. In DNNs, the number of trainable parameters is always far more than that of training samples.
Thus, for this type of over-parameterized model, it is possible to guess that there is a large part of the parameters that will remain unchanged during the whole training phase. More generally, there is a hypothesis that the training trajectory of parameters lies in a subspace constructed by a few irrelevant variables. That is to say, the optimization of millions of parameters can be equivalent to optimization in a tiny subspace. There is also evidence that the gradient of various DNNs will gradually remain in a tiny subspace spanned by a few top eigenvectors of the Hessian  \cite{gur2018gradient}. 

\subsection{Subspace Training}
\par In deep learning, the challenge that the process of training converges very slowly is a thorny obstacle. The slow convergence is caused by the dominating first-order method, \ie{gradient descent-based methods}. This problem can be relieved by second-order methods which utilize the information derived from Hessian matrices. Moreover, the second-order method is not sensitive to the learning rate, so no specific learning rate schedule needs to be designed. However, due to the massive parameters in DNNs, it is a computational burden to calculate Hessian matrices. Some approaches such as Adam \cite{kingma2014adam}, RMSprop \cite{dauphin2015equilibrated} and AdaGrad \cite{duchi2011adaptive} utilize part of  second-order information, like momentum and accumulation information, have already surpassed the performance of conventional gradient-based methods.
\par In order to apply second-order methods such as quasi-Newton method \cite{byrd1994representations} to network training, the straightforward way is to reduce the number of parameters that need to be optimized. In view of the low rank structure discovered in DNNs, it is promising to optimize the whole network in a subspace using quasi-Newton method, without the loss of accuracy. DLDR-based Quasi-Newton method \cite{li2022low} is introduced to save 35\% of training time versus SGD \cite{ruder2016overview}. To be specific, in this algorithm, DLDR is devised to identify the low-dimensional subspace constructed in some important directions which can contribute significantly to the variance of the loss function. It achieves this by sampling the training trajectory and then performing Principal Component Analysis (PCA) to analyse the dominating directions. Then, second-order optimization can be directly executed in this tiny subspace, resulting in fast convergence. 
\subsection{Spatial Redundancy and Temporal Redundancy} \label{Re}
\par While model compression exploits the redundancy in networks to reduce memory and computation complexity, subspace training exploits the redundancy to reduce training time. In other words, the objective of model compression and subspace training is spatial efficiency and temporal efficiency, respectively. Since they both exploit redundancy, we are wondering whether the redundancy they deal with is of the same origin or not. 
\par  We analyse this by performing subspace training on low rank approximated networks to determine if subspace training has a poor performance on compressed networks. If so, it is evidence that the redundancy decreased by model compression is insufficient for subspace training, \ie{the low rank property in time domain disappears}. 

\par
\begin{figure}[!htbp]
    \centering
    \includegraphics[scale=0.6]{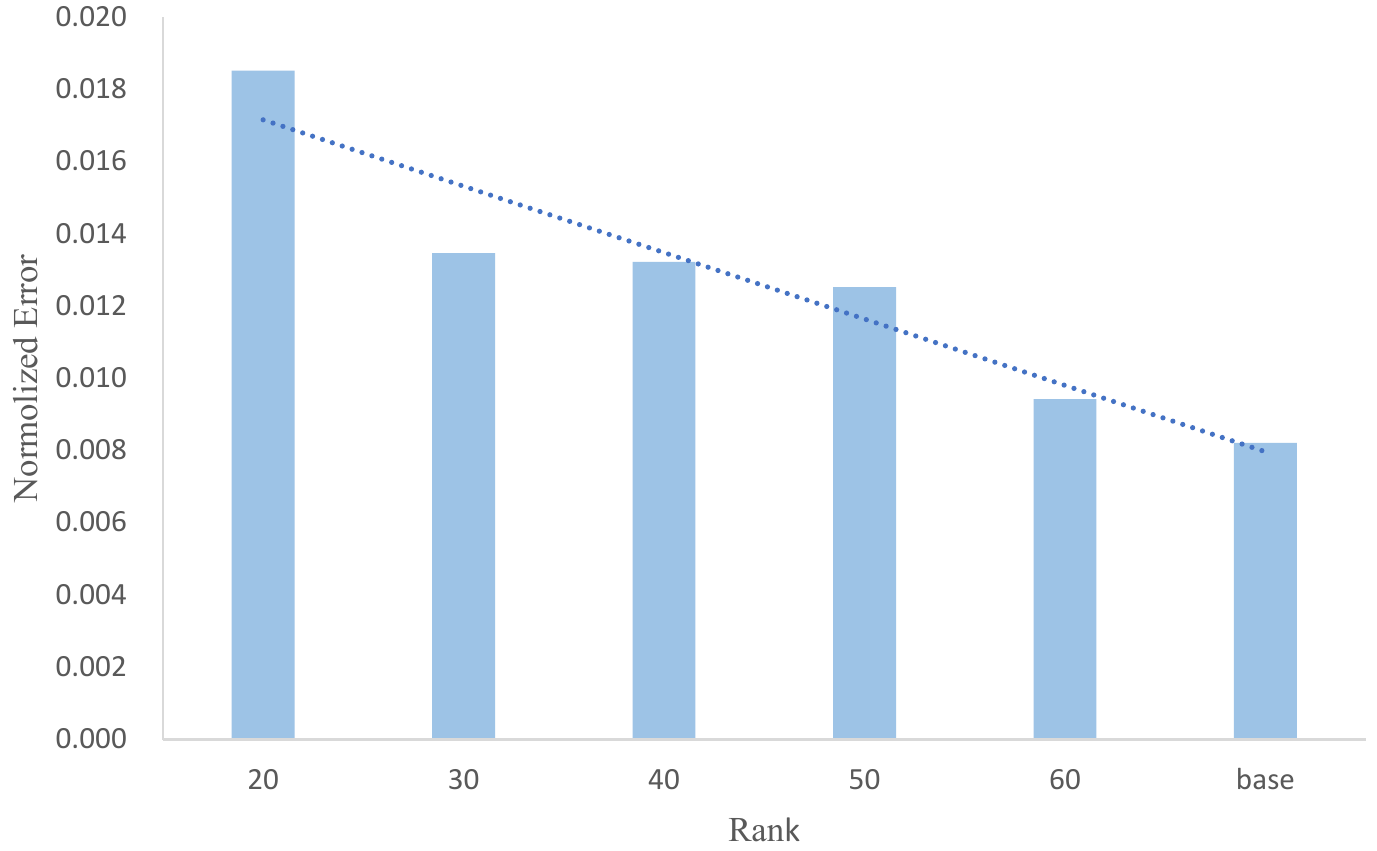}
    \caption{Normalized Error vs rank when projecting SVD-compressed LeNet-300-100 on a 5D subspace. The normalized error is the ratio of $\ell_2$ norm of error between original parameters and projected parameters and $\ell_2$ norm of the original parameters. The rank is in respect to SVD. 'base' is the uncompressed network.}
    \label{rank-error}
\end{figure}

 Here, we perform a simple experiment on LetNet-300-100 with MNIST dataset. LeNet-300-100 contains two hidden fully connected layers with output dimensions 300 and 100, and an output layer with dimension 10. We apply SVD on the first two layers and then fine-tune. We record the training trajectory and establish a 5D subspace by performing PCA. To see if such a tiny subspace is sufficient, we project weights onto this subspace and calculate the normalized approximate error. Fig. \ref{rank-error} shows that as the rank decreases, the normalized error increases almost linearly. It suggests that the higher the compression ratio, the less suitable the subspace with such low fixed dimensionality is. In other words, model compression decreases the redundancy subspace training can exploit. 
\par
\begin{figure}[!htbp]
    \centering
    \includegraphics[scale=0.5]{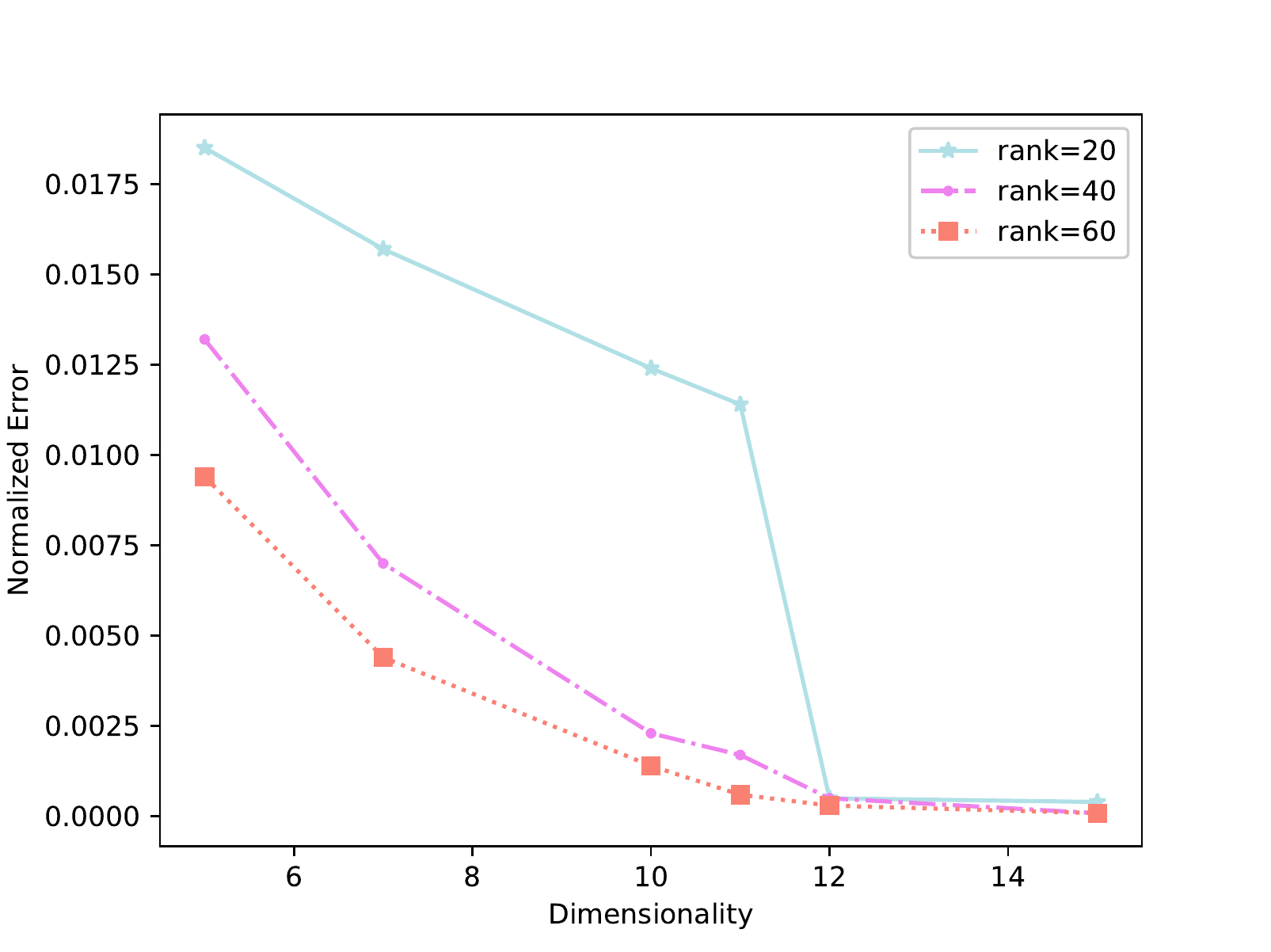}
    \caption{Normalized Error vs dimensionality of subspace under different ranks (different compression extents). The dimensionality of subspace ranges from 5 to 15.}
    \label{D-error}
\end{figure}

 Also, we can figure that after low rank decomposition, a higher-dimensional subspace is in need. As shown in Fig. \ref{D-error}, increasing the dimensionality of subspace has a greater effect on the highly compressed network. Under all the rank settings, normalized error goes to zero when the dimensionality is equal to 12. But there is a sharp descent when the dimensionality is increased from 11 to 12 for rank=20. That is to say, a slight drop in dimensionality is serious for a highly compressed network. When a network is compressed extremely, there is little redundancy in time domain.

\subsection{Make A Balance}
\par Since redundancy exploited by model compression and subspace training are of the same origin, there is a balance between spatial efficiency and temporal efficiency. If we assign most of the redundancy to model compression, we can obtain a compact network and hence achieve spatial efficiency, but little redundancy is left for subspace training. Conversely, if we are in need to train a network quickly, we should promise to assign most of the redundancy to subspace training. 

\par
\begin{figure}[!htbp]
    \centering
    \includegraphics[scale=0.5]{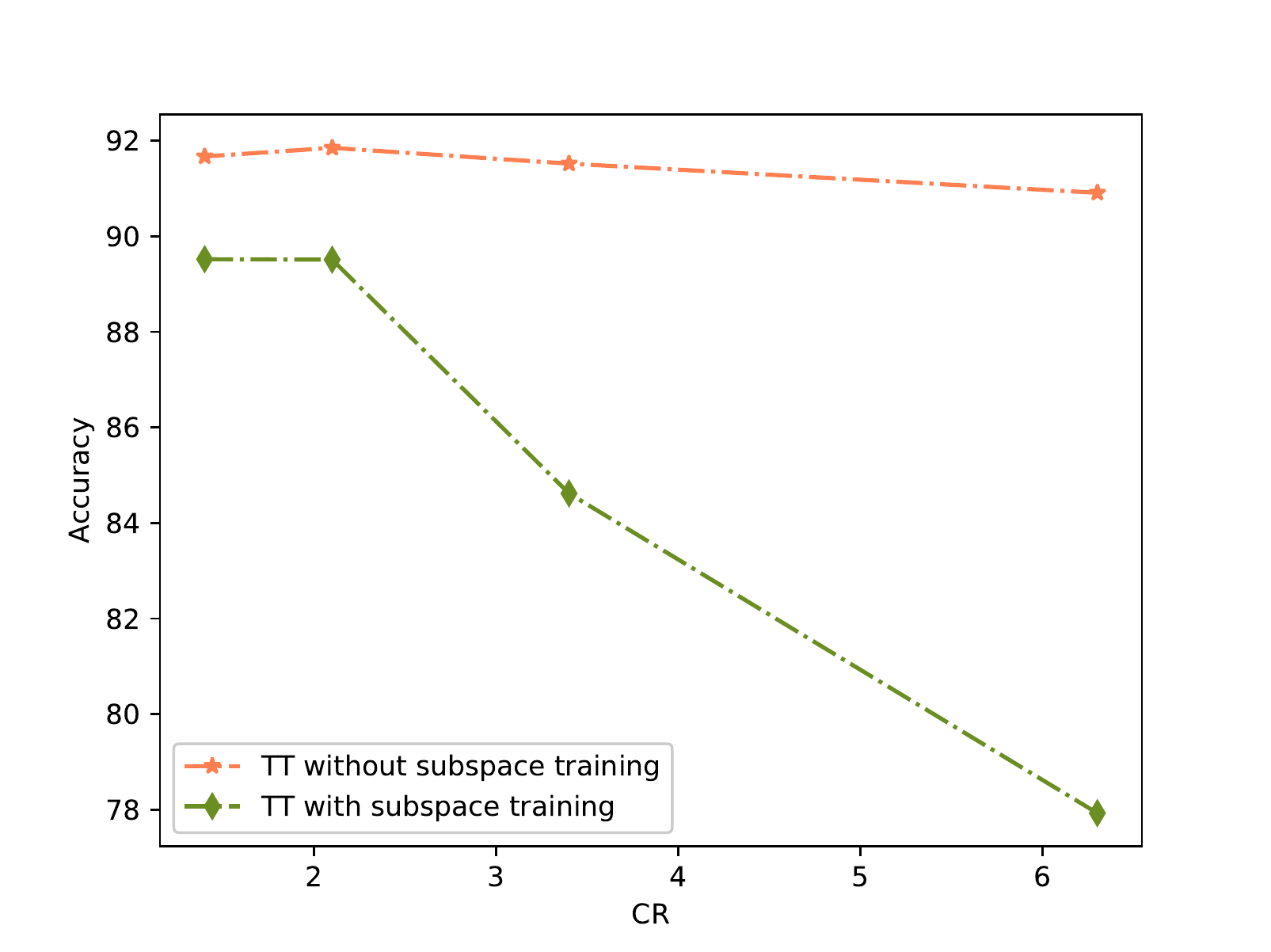}
    \caption{Compare the accuracy degradation when applying subspace training to TT-Nets. CR denotes the compression ratio of model size.}
    \label{accuracy}
\end{figure}

 For model compression, the training of a tensorized neural network (TNN) is much time-consuming than that of the original network. Hence, there is a need for utilizing subspace training to accelerate the training of TNN. Intuitively, for a highly compressed TNN, since there is little redundancy, it is inefficient to train such a TNN in a tiny subspace. Fig. \ref{accuracy} shows the performance of subspace training when applied to TT-based TNNs with various compression regimes. The base network is ResNet32 trained on Cifar10 dataset. All the experiments run 15 epochs (saving 35\% time of SGD method) with Quasi-Newton method and the subspace is fixed to 40D. 
In this figure, the orange line (the case in which TT-Net is trained in normal way) is almost a horizontal line, but the green line (trained in subspace) descends sharply at the time of high compression ratio. It suggests that subspace training can be combined with model compression to achieve spatio-temporal efficiency under a moderate compression regime, but such a tiny space is not suitable for an extremely compressed network.

\par
\begin{figure}
    \centering
    \includegraphics[scale=0.5]{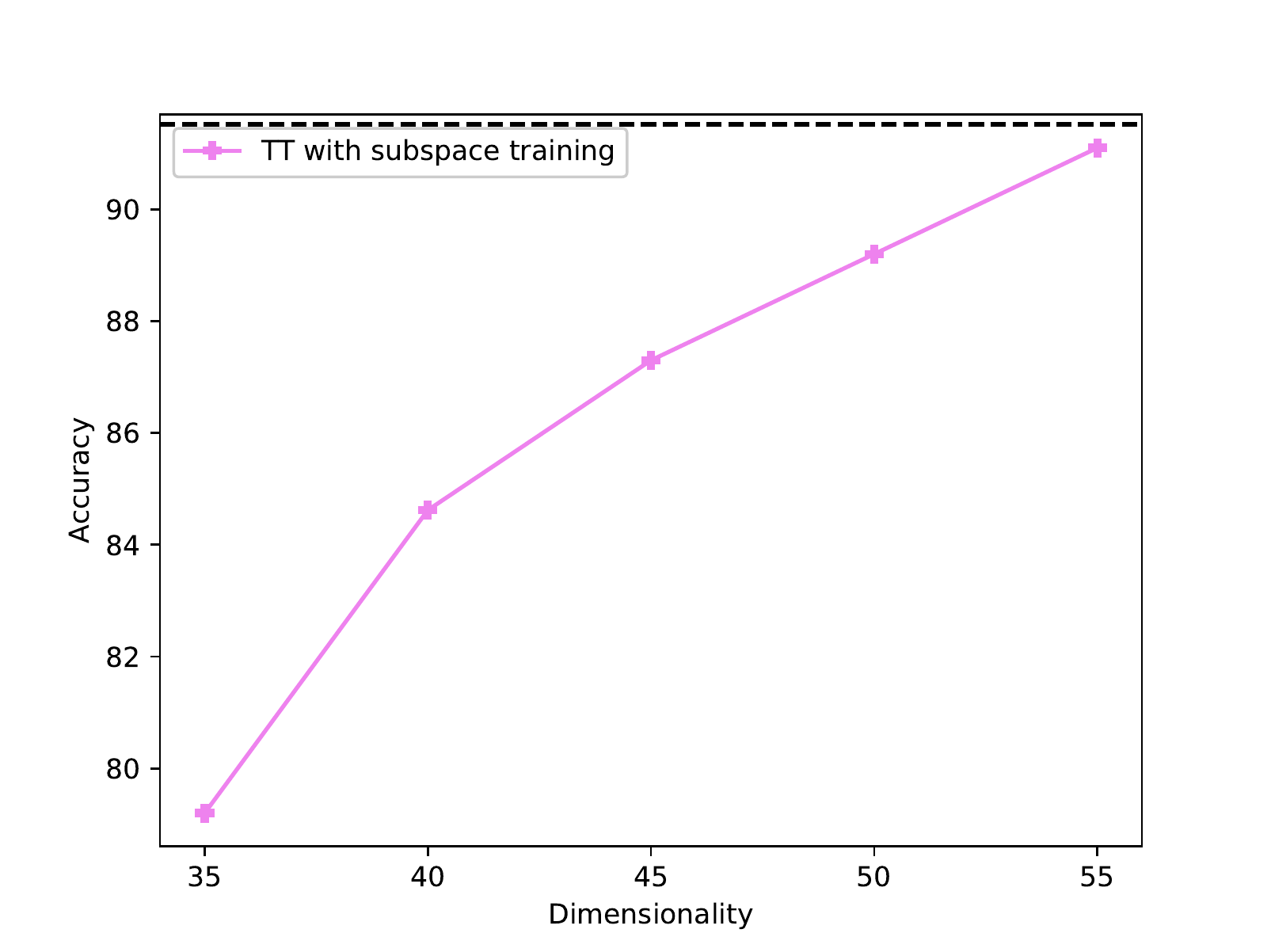}
    \caption{The effect of increasing the dimensionality of subspace on training TT-Nets in subspace. The dashed line represents the accuracy of training TT-Nets in a normal way.}
    \label{TT2}
\end{figure}

 Hence, under an extreme compression regime, it is essential to increase the dimensionality of subspace to relieve the accuracy degradation. But it is infeasible to increase dimensionality blindly, as the number of sampling epochs will also increase, \ie{lessen temporal efficiency}. Fig. \ref{TT2} shows the effect of increasing the dimensionality of subspace for a highly compressed TT-Net. It demonstrates that as the dimensionality of subspace increases, the accuracy degradation of subspace training decreases. When the dimensionality is increased to 55, we can achieve a good accuracy close to the original, but it is worth noting that the total time (time for subspace training and for sampling) is near the normal training time. However, in the case that we want to train a compact TNN quickly and a small drop in accuracy can be tolerated, it is a good choice to train such a network in a moderate subspace.

\section{Conclusion and Future Directions}
\par In this paper, two types of low rank tensor optimization for efficient deep learning are discussed, namely low rank approximation for model compression and subspace training for fast convergence. For low rank approximation, we list various efficient tensor decomposition methods and introduce three types of optimization methods. Since sparsity measure is applied frequently in low rank approximation, we make a comparison among common measures, and experiments show that effective rank can achieve the best accuracy-compression tradeoff. In addition, we investigate how to integrate low rank approximation with other compression techniques. Then, we give a brief introduction to subspace training and analyze that redundancy exploited by subspace training and low rank approximation is of the same origin. Further, we make a discussion on how to combine the two to accelerate the training of tensorized neural networks. 
\par However, up to now, few works focus on integrating more than three types of parameter reduction compression techniques, which is more promising to take maximum advantage of redundancy in networks. Further, it is possible to devise a flexible framework to integrate all kinds of compression techniques. 
\par In practice, low computation complexity is not equivalent to low latency \cite{sze2020evaluate}, and the energy consumed by computation is only a small part of the total energy for inference \cite{horowitz20141,sze2017efficient}. But most works take FLOPs and memory size as benchmarks. That is to say, an advanced algorithm with very low complexity may not be applied to battery-powered mobile devices. Hence, more efforts are needed in decreasing the energy consumption of DNNs.
\par For subspace training, the temporal efficiency is still limited, as the  Quasi-Newton method is still based on the gradient of the original millions of parameters. Direct optimization on several independent variables is still to be studied. In addition, since the sampling procedure occupies most of the training time, there is a need to introduce new techniques to construct subspace with fewer sample epochs. One potential way is to represent all the parameters in tensor format and apply tensor decomposition to better analyze principal components, \ie{higher-order PCA \cite{kolda2009tensor}}.
\section*{Acknowledgment}
The work is supported by the National Natural Science Foundation of China (NSFC, No. 62171088),
Medico-Engineering Cooperation Funds from University
of Electronic Science and Technology of China (No.
ZYGX2021YGLH215).

\ifCLASSOPTIONcaptionsoff
  \newpage
\fi



%


{
\small
\bibliographystyle{ieee_fullname}
\bibliography{ref}

\begin{thebibliography}{100}\itemsep=-1pt

\bibitem{abdul2011accelerating}
Norhamreeza Abdul~Hamid, Nazri Mohd~Nawi, Rozaida Ghazali, and Mohd~Najib
  Mohd~Salleh.
\newblock Accelerating learning performance of back propagation algorithm by
  using adaptive gain together with adaptive momentum and adaptive learning
  rate on classification problems.
\newblock In {\em International Conference on Ubiquitous Computing and
  Multimedia Applications}, pages 559--570. Springer, 2011.

\bibitem{allen2012sparse}
Genevera Allen.
\newblock Sparse higher-order principal components analysis.
\newblock In {\em Artificial Intelligence and Statistics}, pages 27--36. PMLR,
  2012.

\bibitem{alvarez2016learning}
Jose~M Alvarez and Mathieu Salzmann.
\newblock Learning the number of neurons in deep networks.
\newblock {\em Advances in neural information processing systems}, 29, 2016.

\bibitem{alvarez2017compression}
Jose~M Alvarez and Mathieu Salzmann.
\newblock Compression-aware training of deep networks.
\newblock {\em Advances in neural information processing systems}, 30, 2017.

\bibitem{astrid2017cp}
Marcella Astrid and Seung-Ik Lee.
\newblock Cp-decomposition with tensor power method for convolutional neural
  networks compression.
\newblock In {\em 2017 IEEE International Conference on Big Data and Smart
  Computing (BigComp)}, pages 115--118. IEEE, 2017.

\bibitem{avron2012efficient}
Haim Avron, Satyen Kale, Shiva Kasiviswanathan, and Vikas Sindhwani.
\newblock Efficient and practical stochastic subgradient descent for nuclear
  norm regularization.
\newblock {\em arXiv preprint arXiv:1206.6384}, 2012.

\bibitem{bazerque2013rank}
Juan~Andr{\'e}s Bazerque, Gonzalo Mateos, and Georgios~B Giannakis.
\newblock Rank regularization and bayesian inference for tensor completion and
  extrapolation.
\newblock {\em IEEE transactions on signal processing}, 61(22):5689--5703,
  2013.

\bibitem{blalock2020state}
Davis Blalock, Jose~Javier Gonzalez~Ortiz, Jonathan Frankle, and John Guttag.
\newblock What is the state of neural network pruning?
\newblock {\em Proceedings of machine learning and systems}, 2:129--146, 2020.

\bibitem{bogdan2013statistical}
Malgorzata Bogdan, Ewout van~den Berg, Weijie Su, and Emmanuel Candes.
\newblock Statistical estimation and testing via the sorted l1 norm.
\newblock {\em arXiv preprint arXiv:1310.1969}, 2013.

\bibitem{byrd1994representations}
Richard~H Byrd, Jorge Nocedal, and Robert~B Schnabel.
\newblock Representations of quasi-newton matrices and their use in limited
  memory methods.
\newblock {\em Mathematical Programming}, 63(1):129--156, 1994.

\bibitem{cai2010singular}
Jian-Feng Cai, Emmanuel~J Cand{\`e}s, and Zuowei Shen.
\newblock A singular value thresholding algorithm for matrix completion.
\newblock {\em SIAM Journal on optimization}, 20(4):1956--1982, 2010.

\bibitem{carreira2018learning}
Miguel~A Carreira-Perpin{\'a}n and Yerlan Idelbayev.
\newblock “learning-compression” algorithms for neural net pruning.
\newblock In {\em Proceedings of the IEEE Conference on Computer Vision and
  Pattern Recognition}, pages 8532--8541, 2018.

\bibitem{chaudhuri2016computational}
Rishidev Chaudhuri and Ila Fiete.
\newblock Computational principles of memory.
\newblock {\em Nature neuroscience}, 19(3):394--403, 2016.

\bibitem{chen2021efficient}
Chunyun Chen, Zhe Wang, Xiaowei Chen, Jie Lin, and Mohamed M~Sabry Aly.
\newblock Efficient tunstall decoder for deep neural network compression.
\newblock In {\em 2021 58th ACM/IEEE Design Automation Conference (DAC)}, pages
  1021--1026. IEEE, 2021.

\bibitem{chen2018shallowing}
Shi Chen and Qi Zhao.
\newblock Shallowing deep networks: Layer-wise pruning based on feature
  representations.
\newblock {\em IEEE transactions on pattern analysis and machine intelligence},
  41(12):3048--3056, 2018.

\bibitem{chen2015compressing}
Wenlin Chen, James Wilson, Stephen Tyree, Kilian Weinberger, and Yixin Chen.
\newblock Compressing neural networks with the hashing trick.
\newblock In {\em International conference on machine learning}, pages
  2285--2294. PMLR, 2015.

\bibitem{chen2020deep}
Zhen Chen, Zhibo Chen, Jianxin Lin, Sen Liu, and Weiping Li.
\newblock Deep neural network acceleration based on low-rank approximated
  channel pruning.
\newblock {\em IEEE Transactions on Circuits and Systems I: Regular Papers},
  67(4):1232--1244, 2020.

\bibitem{cheng2007survey}
Daizhan Cheng, Hongsheng Qi, and Ancheng Xue.
\newblock A survey on semi-tensor product of matrices.
\newblock {\em Journal of Systems Science and Complexity}, 20(2):304--322,
  2007.

\bibitem{cheng2020novel}
Zhiyu Cheng, Baopu Li, Yanwen Fan, and Yingze Bao.
\newblock A novel rank selection scheme in tensor ring decomposition based on
  reinforcement learning for deep neural networks.
\newblock In {\em ICASSP 2020-2020 IEEE International Conference on Acoustics,
  Speech and Signal Processing (ICASSP)}, pages 3292--3296. IEEE, 2020.

\bibitem{choi2020universal}
Yoojin Choi, Mostafa El-Khamy, and Jungwon Lee.
\newblock Universal deep neural network compression.
\newblock {\em IEEE Journal of Selected Topics in Signal Processing},
  14(4):715--726, 2020.

\bibitem{choudhary2020comprehensive}
Tejalal Choudhary, Vipul Mishra, Anurag Goswami, and Jagannathan Sarangapani.
\newblock A comprehensive survey on model compression and acceleration.
\newblock {\em Artificial Intelligence Review}, 53(7):5113--5155, 2020.

\bibitem{constantine2015exploiting}
Paul~G Constantine, Michael Emory, Johan Larsson, and Gianluca Iaccarino.
\newblock Exploiting active subspaces to quantify uncertainty in the numerical
  simulation of the hyshot ii scramjet.
\newblock {\em Journal of Computational Physics}, 302:1--20, 2015.

\bibitem{cosson2022gradient}
Romain Cosson, Ali Jadbabaie, Anuran Makur, Amirhossein Reisizadeh, and
  Devavrat Shah.
\newblock Gradient descent for low-rank functions.
\newblock {\em arXiv preprint arXiv:2206.08257}, 2022.

\bibitem{dalton1920measurement}
Hugh Dalton.
\newblock The measurement of the inequality of incomes.
\newblock {\em The Economic Journal}, 30(119):348--361, 1920.

\bibitem{dauphin2015equilibrated}
Yann Dauphin, Harm De~Vries, and Yoshua Bengio.
\newblock Equilibrated adaptive learning rates for non-convex optimization.
\newblock {\em Advances in neural information processing systems}, 28, 2015.

\bibitem{de2008decompositions}
Lieven De~Lathauwer.
\newblock Decompositions of a higher-order tensor in block terms—part ii:
  Definitions and uniqueness.
\newblock {\em SIAM Journal on Matrix Analysis and Applications},
  30(3):1033--1066, 2008.

\bibitem{deb2002fast}
Kalyanmoy Deb, Amrit Pratap, Sameer Agarwal, and TAMT Meyarivan.
\newblock A fast and elitist multiobjective genetic algorithm: Nsga-ii.
\newblock {\em IEEE transactions on evolutionary computation}, 6(2):182--197,
  2002.

\bibitem{deng2020model}
Lei Deng, Guoqi Li, Song Han, Luping Shi, and Yuan Xie.
\newblock Model compression and hardware acceleration for neural networks: A
  comprehensive survey.
\newblock {\em Proceedings of the IEEE}, 108(4):485--532, 2020.

\bibitem{denil2013predicting}
Misha Denil, Babak Shakibi, Laurent Dinh, Marc'Aurelio Ranzato, and Nando
  De~Freitas.
\newblock Predicting parameters in deep learning.
\newblock {\em Advances in neural information processing systems}, 26, 2013.

\bibitem{denton2014exploiting}
Emily~L Denton, Wojciech Zaremba, Joan Bruna, Yann LeCun, and Rob Fergus.
\newblock Exploiting linear structure within convolutional networks for
  efficient evaluation.
\newblock {\em Advances in neural information processing systems}, 27, 2014.

\bibitem{donoho1989projection}
David~L Donoho and Iain~M Johnstone.
\newblock Projection-based approximation and a duality with kernel methods.
\newblock {\em The Annals of Statistics}, pages 58--106, 1989.

\bibitem{duchi2011adaptive}
John Duchi, Elad Hazan, and Yoram Singer.
\newblock Adaptive subgradient methods for online learning and stochastic
  optimization.
\newblock {\em Journal of machine learning research}, 12(7), 2011.

\bibitem{effros2002universal}
Michelle Effros, Karthik Visweswariah, Sanjeev~R Kulkarni, and Sergio
  Verd{\'u}.
\newblock Universal lossless source coding with the burrows wheeler transform.
\newblock {\em IEEE Transactions on Information Theory}, 48(5):1061--1081,
  2002.

\bibitem{eo2023effective}
Moonjung Eo, Suhyun Kang, and Wonjong Rhee.
\newblock An effective low-rank compression with a joint rank selection
  followed by a compression-friendly training.
\newblock {\em Neural Networks}, 161:165--177, 2023.

\bibitem{espig2011optimization}
Mike Espig, Wolfgang Hackbusch, Stefan Handschuh, and Reinhold Schneider.
\newblock Optimization problems in contracted tensor networks.
\newblock {\em Computing and visualization in science}, 14(6):271--285, 2011.

\bibitem{faisal2008noise}
A~Aldo Faisal, Luc~PJ Selen, and Daniel~M Wolpert.
\newblock Noise in the nervous system.
\newblock {\em Nature reviews neuroscience}, 9(4):292--303, 2008.

\bibitem{garipov2016ultimate}
Timur Garipov, Dmitry Podoprikhin, Alexander Novikov, and Dmitry Vetrov.
\newblock Ultimate tensorization: compressing convolutional and fc layers
  alike.
\newblock {\em arXiv preprint arXiv:1611.03214}, 2016.

\bibitem{gholami2021survey}
Amir Gholami, Sehoon Kim, Zhen Dong, Zhewei Yao, Michael~W Mahoney, and Kurt
  Keutzer.
\newblock A survey of quantization methods for efficient neural network
  inference.
\newblock {\em arXiv preprint arXiv:2103.13630}, 2021.

\bibitem{glorot2010understanding}
Xavier Glorot and Yoshua Bengio.
\newblock Understanding the difficulty of training deep feedforward neural
  networks.
\newblock In {\em Proceedings of the thirteenth international conference on
  artificial intelligence and statistics}, pages 249--256. JMLR Workshop and
  Conference Proceedings, 2010.

\bibitem{gong2014compressing}
Yunchao Gong, Liu Liu, Ming Yang, and Lubomir Bourdev.
\newblock Compressing deep convolutional networks using vector quantization.
\newblock {\em arXiv preprint arXiv:1412.6115}, 2014.

\bibitem{goyal2019compression}
Saurabh Goyal, Anamitra~Roy Choudhury, and Vivek Sharma.
\newblock Compression of deep neural networks by combining pruning and low rank
  decomposition.
\newblock In {\em 2019 IEEE International Parallel and Distributed Processing
  Symposium Workshops (IPDPSW)}, pages 952--958. IEEE, 2019.

\bibitem{grasedyck2010hierarchical}
Lars Grasedyck.
\newblock Hierarchical singular value decomposition of tensors.
\newblock {\em SIAM journal on matrix analysis and applications},
  31(4):2029--2054, 2010.

\bibitem{graves2013speech}
Alex Graves, Abdel-rahman Mohamed, and Geoffrey Hinton.
\newblock Speech recognition with deep recurrent neural networks.
\newblock In {\em 2013 IEEE international conference on acoustics, speech and
  signal processing}, pages 6645--6649. Ieee, 2013.

\bibitem{guhaniyogi2017bayesian}
Rajarshi Guhaniyogi, Shaan Qamar, and David~B Dunson.
\newblock Bayesian tensor regression.
\newblock {\em The Journal of Machine Learning Research}, 18(1):2733--2763,
  2017.

\bibitem{gur2018gradient}
Guy Gur-Ari, Daniel~A Roberts, and Ethan Dyer.
\newblock Gradient descent happens in a tiny subspace.
\newblock {\em arXiv preprint arXiv:1812.04754}, 2018.

\bibitem{hameed2022convolutional}
Marawan Gamal~Abdel Hameed, Marzieh~S Tahaei, Ali Mosleh, and Vahid~Partovi
  Nia.
\newblock Convolutional neural network compression through generalized
  kronecker product decomposition.
\newblock In {\em Proceedings of the AAAI Conference on Artificial
  Intelligence}, pages 771--779, 2022.

\bibitem{han2016eie}
Song Han, Xingyu Liu, Huizi Mao, Jing Pu, Ardavan Pedram, Mark~A Horowitz, and
  William~J Dally.
\newblock Eie: Efficient inference engine on compressed deep neural network.
\newblock {\em ACM SIGARCH Computer Architecture News}, 44(3):243--254, 2016.

\bibitem{han2015deep}
Song Han, Huizi Mao, and William~J Dally.
\newblock Deep compression: Compressing deep neural networks with pruning,
  trained quantization and huffman coding.
\newblock {\em arXiv preprint arXiv:1510.00149}, 2015.

\bibitem{han2015learning}
Song Han, Jeff Pool, John Tran, and William Dally.
\newblock Learning both weights and connections for efficient neural network.
\newblock {\em Advances in neural information processing systems}, 28, 2015.

\bibitem{harshman2004problem}
Richard~A Harshman.
\newblock The problem and nature of degenerate solutions or decompositions of
  3-way arrays.
\newblock In {\em Talk at the Tensor Decompositions Workshop, Palo Alto, CA,
  American Institute of Mathematics}, 2004.

\bibitem{hawkins2021bayesian}
Cole Hawkins and Zheng Zhang.
\newblock Bayesian tensorized neural networks with automatic rank selection.
\newblock {\em Neurocomputing}, 453:172--180, 2021.

\bibitem{hinton2015distilling}
Geoffrey Hinton, Oriol Vinyals, Jeff Dean, et~al.
\newblock Distilling the knowledge in a neural network.
\newblock {\em arXiv preprint arXiv:1503.02531}, 2(7), 2015.

\bibitem{hinton2012improving}
Geoffrey~E Hinton, Nitish Srivastava, Alex Krizhevsky, Ilya Sutskever, and
  Ruslan~R Salakhutdinov.
\newblock Improving neural networks by preventing co-adaptation of feature
  detectors.
\newblock {\em arXiv preprint arXiv:1207.0580}, 2012.

\bibitem{horowitz20141}
Mark Horowitz.
\newblock 1.1 computing's energy problem (and what we can do about it).
\newblock In {\em 2014 IEEE International Solid-State Circuits Conference
  Digest of Technical Papers (ISSCC)}, pages 10--14. IEEE, 2014.

\bibitem{hoyer2004non}
Patrik~O Hoyer.
\newblock Non-negative matrix factorization with sparseness constraints.
\newblock {\em Journal of machine learning research}, 5(9), 2004.

\bibitem{huang2020robust}
Huyan Huang, Yipeng Liu, Zhen Long, and Ce Zhu.
\newblock Robust low-rank tensor ring completion.
\newblock {\em IEEE Transactions on Computational Imaging}, 6:1117--1126, 2020.

\bibitem{huang2011learning}
Junzhou Huang, Tong Zhang, and Dimitris Metaxas.
\newblock Learning with structured sparsity.
\newblock {\em Journal of Machine Learning Research}, 12(11), 2011.

\bibitem{huang2018learning}
Qiangui Huang, Kevin Zhou, Suya You, and Ulrich Neumann.
\newblock Learning to prune filters in convolutional neural networks.
\newblock In {\em 2018 IEEE Winter Conference on Applications of Computer
  Vision (WACV)}, pages 709--718. IEEE, 2018.

\bibitem{huang2015two}
Xiaolin Huang, Yipeng Liu, Lei Shi, Sabine Van~Huffel, and Johan~AK Suykens.
\newblock Two-level $\ell_1$ minimization for compressed sensing.
\newblock {\em Signal Processing}, 108:459--475, 2015.

\bibitem{hurley2005parameterized}
Niall Hurley, Scott Rickard, and Paul Curran.
\newblock Parameterized lifting for sparse signal representations using the
  gini index.
\newblock {\em Signal Processing with Adaptative Sparse Structured
  Representations (SPARS05), Rennes, France}, 2005.

\bibitem{idelbayev2020low}
Yerlan Idelbayev and Miguel~A Carreira-Perpin{\'a}n.
\newblock Low-rank compression of neural nets: Learning the rank of each layer.
\newblock In {\em Proceedings of the IEEE/CVF Conference on Computer Vision and
  Pattern Recognition}, pages 8049--8059, 2020.

\bibitem{jaderberg2014speeding}
Max Jaderberg, Andrea Vedaldi, and Andrew Zisserman.
\newblock Speeding up convolutional neural networks with low rank expansions.
\newblock {\em arXiv preprint arXiv:1405.3866}, 2014.

\bibitem{jiang2017exploiting}
Yu-Gang Jiang, Zuxuan Wu, Jun Wang, Xiangyang Xue, and Shih-Fu Chang.
\newblock Exploiting feature and class relationships in video categorization
  with regularized deep neural networks.
\newblock {\em IEEE transactions on pattern analysis and machine intelligence},
  40(2):352--364, 2017.

\bibitem{khaw2017discrete}
Mel~Win Khaw, Luminita Stevens, and Michael Woodford.
\newblock Discrete adjustment to a changing environment: Experimental evidence.
\newblock {\em Journal of Monetary Economics}, 91:88--103, 2017.

\bibitem{kim2015compression}
Yong-Deok Kim, Eunhyeok Park, Sungjoo Yoo, Taelim Choi, Lu Yang, and Dongjun
  Shin.
\newblock Compression of deep convolutional neural networks for fast and low
  power mobile applications.
\newblock {\em arXiv preprint arXiv:1511.06530}, 2015.

\bibitem{kingma2014adam}
Diederik~P Kingma and Jimmy Ba.
\newblock Adam: A method for stochastic optimization.
\newblock {\em arXiv preprint arXiv:1412.6980}, 2014.

\bibitem{kolda2009tensor}
Tamara~G Kolda and Brett~W Bader.
\newblock Tensor decompositions and applications.
\newblock {\em SIAM review}, 51(3):455--500, 2009.

\bibitem{kozyrskiy2020cnn}
Nikolay Kozyrskiy and Anh-Huy Phan.
\newblock Cnn acceleration by low-rank approximation with quantized factors.
\newblock {\em arXiv preprint arXiv:2006.08878}, 2020.

\bibitem{krijnen2008non}
Wim~P Krijnen, Theo~K Dijkstra, and Alwin Stegeman.
\newblock On the non-existence of optimal solutions and the occurrence of
  “degeneracy” in the candecomp/parafac model.
\newblock {\em Psychometrika}, 73(3):431--439, 2008.

\bibitem{krizhevsky2012imagenet}
Alex Krizhevsky, Ilya Sutskever, and Geoffrey~E Hinton.
\newblock Imagenet classification with deep convolutional neural networks.
\newblock {\em Advances in neural information processing systems}, 25, 2012.

\bibitem{kumar2022vision}
Ankur Kumar.
\newblock Vision transformer compression with structured pruning and low rank
  approximation.
\newblock {\em arXiv preprint arXiv:2203.13444}, 2022.

\bibitem{kuzmin2022quantized}
Andrey Kuzmin, Mart van Baalen, Markus Nagel, and Arash Behboodi.
\newblock Quantized sparse weight decomposition for neural network compression.
\newblock {\em arXiv preprint arXiv:2207.11048}, 2022.

\bibitem{lane2015early}
Nicholas~D Lane, Sourav Bhattacharya, Petko Georgiev, Claudio Forlivesi, and
  Fahim Kawsar.
\newblock An early resource characterization of deep learning on wearables,
  smartphones and internet-of-things devices.
\newblock In {\em Proceedings of the 2015 international workshop on internet of
  things towards applications}, pages 7--12, 2015.

\bibitem{latimer2015single}
Kenneth~W Latimer, Jacob~L Yates, Miriam~LR Meister, Alexander~C Huk, and
  Jonathan~W Pillow.
\newblock Single-trial spike trains in parietal cortex reveal discrete steps
  during decision-making.
\newblock {\em Science}, 349(6244):184--187, 2015.

\bibitem{lebedev2014speeding}
Vadim Lebedev, Yaroslav Ganin, Maksim Rakhuba, Ivan Oseledets, and Victor
  Lempitsky.
\newblock Speeding-up convolutional neural networks using fine-tuned
  cp-decomposition.
\newblock {\em arXiv preprint arXiv:1412.6553}, 2014.

\bibitem{lebedev2016fast}
Vadim Lebedev and Victor Lempitsky.
\newblock Fast convnets using group-wise brain damage.
\newblock In {\em Proceedings of the IEEE Conference on Computer Vision and
  Pattern Recognition}, pages 2554--2564, 2016.

\bibitem{lecun2015deep}
Yann LeCun, Yoshua Bengio, and Geoffrey Hinton.
\newblock Deep learning.
\newblock {\em nature}, 521(7553):436--444, 2015.

\bibitem{lee2021qttnet}
Donghyun Lee, Dingheng Wang, Yukuan Yang, Lei Deng, Guangshe Zhao, and Guoqi
  Li.
\newblock Qttnet: Quantized tensor train neural networks for 3d object and
  video recognition.
\newblock {\em Neural Networks}, 141:420--432, 2021.

\bibitem{li2022heuristic}
Nannan Li, Yu Pan, Yaran Chen, Zixiang Ding, Dongbin Zhao, and Zenglin Xu.
\newblock Heuristic rank selection with progressively searching tensor ring
  network.
\newblock {\em Complex \& Intelligent Systems}, 8(2):771--785, 2022.

\bibitem{li2020few}
Tianhong Li, Jianguo Li, Zhuang Liu, and Changshui Zhang.
\newblock Few sample knowledge distillation for efficient network compression.
\newblock In {\em Proceedings of the IEEE/CVF Conference on Computer Vision and
  Pattern Recognition}, pages 14639--14647, 2020.

\bibitem{li2022low}
Tao Li, Lei Tan, Zhehao Huang, Qinghua Tao, Yipeng Liu, and Xiaolin Huang.
\newblock Low dimensional trajectory hypothesis is true: Dnns can be trained in
  tiny subspaces.
\newblock {\em IEEE Transactions on Pattern Analysis and Machine Intelligence},
  2022.

\bibitem{li2019learning}
Yawei Li, Shuhang Gu, Luc~Van Gool, and Radu Timofte.
\newblock Learning filter basis for convolutional neural network compression.
\newblock In {\em Proceedings of the IEEE/CVF International Conference on
  Computer Vision}, pages 5623--5632, 2019.

\bibitem{liebenwein2021compressing}
Lucas Liebenwein, Alaa Maalouf, Dan Feldman, and Daniela Rus.
\newblock Compressing neural networks: Towards determining the optimal
  layer-wise decomposition.
\newblock {\em Advances in Neural Information Processing Systems},
  34:5328--5344, 2021.

\bibitem{lin2016fixed}
Darryl Lin, Sachin Talathi, and Sreekanth Annapureddy.
\newblock Fixed point quantization of deep convolutional networks.
\newblock In {\em International conference on machine learning}, pages
  2849--2858. PMLR, 2016.

\bibitem{lin2018holistic}
Shaohui Lin, Rongrong Ji, Chao Chen, Dacheng Tao, and Jiebo Luo.
\newblock Holistic cnn compression via low-rank decomposition with knowledge
  transfer.
\newblock {\em IEEE transactions on pattern analysis and machine intelligence},
  41(12):2889--2905, 2018.

\bibitem{liu2015sparse}
Baoyuan Liu, Min Wang, Hassan Foroosh, Marshall Tappen, and Marianna Pensky.
\newblock Sparse convolutional neural networks.
\newblock In {\em Proceedings of the IEEE conference on computer vision and
  pattern recognition}, pages 806--814, 2015.

\bibitem{liu2020smooth}
Jiani Liu, Ce Zhu, and Yipeng Liu.
\newblock Smooth compact tensor ring regression.
\newblock {\em IEEE Transactions on Knowledge and Data Engineering},
  34(9):4439--4452, 2020.

\bibitem{liu2021tensor}
Jiani Liu, Ce Zhu, Zhen Long, Yipeng Liu, et~al.
\newblock Tensor regression.
\newblock {\em Foundations and Trends{\textregistered} in Machine Learning},
  14(4):379--565, 2021.

\bibitem{liu2013efficient}
Xing Liu, Mikhail Smelyanskiy, Edmond Chow, and Pradeep Dubey.
\newblock Efficient sparse matrix-vector multiplication on x86-based many-core
  processors.
\newblock In {\em Proceedings of the 27th international ACM conference on
  International conference on supercomputing}, pages 273--282, 2013.

\bibitem{liu2021tensors}
Yipeng Liu.
\newblock {\em Tensors for Data Processing: Theory, Methods, and Applications}.
\newblock Academic Press, 2021.

\bibitem{liu2013multi}
Yipeng Liu, Maarten De~Vos, Ivan Gligorijevic, Vladimir Matic, Yuqian Li, and
  Sabine Van~Huffel.
\newblock Multi-structural signal recovery for biomedical compressive sensing.
\newblock {\em IEEE Transactions on Biomedical Engineering}, 60(10):2794--2805,
  2013.

\bibitem{liu2022tensor}
Yipeng Liu, Jiani Liu, Zhen Long, and Ce Zhu.
\newblock Tensor decomposition in deep networks.
\newblock In {\em Tensor Computation for Data Analysis}, pages 241--263.
  Springer, 2022.

\bibitem{liu2020low}
Yipeng Liu, Jiani Liu, and Ce Zhu.
\newblock Low-rank tensor train coefficient array estimation for
  tensor-on-tensor regression.
\newblock {\em IEEE transactions on neural networks and learning systems},
  31(12):5402--5411, 2020.

\bibitem{liu2019low}
Yipeng Liu, Zhen Long, Huyan Huang, and Ce Zhu.
\newblock Low cp rank and tucker rank tensor completion for estimating missing
  components in image data.
\newblock {\em IEEE Transactions on Circuits and Systems for Video Technology},
  30(4):944--954, 2019.

\bibitem{liu2018image}
Yipeng Liu, Zhen Long, and Ce Zhu.
\newblock Image completion using low tensor tree rank and total variation
  minimization.
\newblock {\em IEEE Transactions on Multimedia}, 21(2):338--350, 2018.

\bibitem{logan1975optimal}
Benjamin~F Logan and Larry~A Shepp.
\newblock Optimal reconstruction of a function from its projections.
\newblock {\em Duke mathematical journal}, 42(4):645--659, 1975.

\bibitem{long2021bayesian}
Zhen Long, Ce Zhu, Jiani Liu, and Yipeng Liu.
\newblock Bayesian low rank tensor ring for image recovery.
\newblock {\em IEEE Transactions on Image Processing}, 30:3568--3580, 2021.

\bibitem{lorenz1905methods}
Max~O Lorenz.
\newblock Methods of measuring the concentration of wealth.
\newblock {\em Publications of the American statistical association},
  9(70):209--219, 1905.

\bibitem{luo2017thinet}
Jian-Hao Luo, Jianxin Wu, and Weiyao Lin.
\newblock Thinet: A filter level pruning method for deep neural network
  compression.
\newblock In {\em Proceedings of the IEEE international conference on computer
  vision}, pages 5058--5066, 2017.

\bibitem{mitchell1994slowly}
Ben~C Mitchell and Donald~S Burdick.
\newblock Slowly converging parafac sequences: swamps and two-factor
  degeneracies.
\newblock {\em Journal of Chemometrics}, 8(2):155--168, 1994.

\bibitem{molchanov2016pruning}
Pavlo Molchanov, Stephen Tyree, Tero Karras, Timo Aila, and Jan Kautz.
\newblock Pruning convolutional neural networks for resource efficient
  inference.
\newblock {\em arXiv preprint arXiv:1611.06440}, 2016.

\bibitem{nagel2021white}
Markus Nagel, Marios Fournarakis, Rana~Ali Amjad, Yelysei Bondarenko, Mart van
  Baalen, and Tijmen Blankevoort.
\newblock A white paper on neural network quantization.
\newblock {\em arXiv preprint arXiv:2106.08295}, 2021.

\bibitem{nakajima2013global}
Shinichi Nakajima, Masashi Sugiyama, S~Derin Babacan, and Ryota Tomioka.
\newblock Global analytic solution of fully-observed variational bayesian
  matrix factorization.
\newblock {\em The Journal of Machine Learning Research}, 14(1):1--37, 2013.

\bibitem{nekooei2022compression}
Amirreza Nekooei and Saeed Safari.
\newblock Compression of deep neural networks based on quantized tensor
  decomposition to implement on reconfigurable hardware platforms.
\newblock {\em Neural Networks}, 150:350--363, 2022.

\bibitem{novikov2015tensorizing}
Alexander Novikov, Dmitrii Podoprikhin, Anton Osokin, and Dmitry~P Vetrov.
\newblock Tensorizing neural networks.
\newblock {\em Advances in neural information processing systems}, 28, 2015.

\bibitem{obukhov2020t}
Anton Obukhov, Maxim Rakhuba, Stamatios Georgoulis, Menelaos Kanakis, Dengxin
  Dai, and Luc Van~Gool.
\newblock T-basis: a compact representation for neural networks.
\newblock In {\em International Conference on Machine Learning}, pages
  7392--7404. PMLR, 2020.

\bibitem{osawa2017evaluating}
Kazuki Osawa and Rio Yokota.
\newblock Evaluating the compression efficiency of the filters in convolutional
  neural networks.
\newblock In {\em International Conference on Artificial Neural Networks},
  pages 459--466. Springer, 2017.

\bibitem{oseledets2011tensor}
Ivan~V Oseledets.
\newblock Tensor-train decomposition.
\newblock {\em SIAM Journal on Scientific Computing}, 33(5):2295--2317, 2011.

\bibitem{pan2022unified}
Yu Pan, Zeyong Su, Ao Liu, Wang Jingquan, Nannan Li, and Zenglin Xu.
\newblock A unified weight initialization paradigm for tensorial convolutional
  neural networks.
\newblock In {\em International Conference on Machine Learning}, pages
  17238--17257. PMLR, 2022.

\bibitem{phan2022train}
Anh-Huy Phan, Konstantin Sobolev, Dmitry Ermilov, Igor Vorona, Nikolay
  Kozyrskiy, Petr Tichavsky, and Andrzej Cichocki.
\newblock How to train unstable looped tensor network.
\newblock {\em arXiv preprint arXiv:2203.02617}, 2022.

\bibitem{phan2020stable}
Anh-Huy Phan, Konstantin Sobolev, Konstantin Sozykin, Dmitry Ermilov, Julia
  Gusak, Petr Tichavsk{\`y}, Valeriy Glukhov, Ivan Oseledets, and Andrzej
  Cichocki.
\newblock Stable low-rank tensor decomposition for compression of convolutional
  neural network.
\newblock In {\em European Conference on Computer Vision}, pages 522--539.
  Springer, 2020.

\bibitem{rai2014scalable}
Piyush Rai, Yingjian Wang, Shengbo Guo, Gary Chen, David Dunson, and Lawrence
  Carin.
\newblock Scalable bayesian low-rank decomposition of incomplete multiway
  tensors.
\newblock In {\em International Conference on Machine Learning}, pages
  1800--1808. PMLR, 2014.

\bibitem{recanatesi2019dimensionality}
Stefano Recanatesi, Matthew Farrell, Madhu Advani, Timothy Moore, Guillaume
  Lajoie, and Eric Shea-Brown.
\newblock Dimensionality compression and expansion in deep neural networks.
\newblock {\em arXiv preprint arXiv:1906.00443}, 2019.

\bibitem{reeves1993modern}
Colin~R Reeves.
\newblock {\em Modern heuristic techniques for combinatorial problems}.
\newblock John Wiley \& Sons, Inc., 1993.

\bibitem{rickard2006sparse}
Scott Rickard.
\newblock Sparse sources are separated sources.
\newblock In {\em 2006 14th European signal processing conference}, pages 1--5.
  IEEE, 2006.

\bibitem{roy2007effective}
Olivier Roy and Martin Vetterli.
\newblock The effective rank: A measure of effective dimensionality.
\newblock In {\em 2007 15th European signal processing conference}, pages
  606--610. IEEE, 2007.

\bibitem{ruan2020edp}
Xiaofeng Ruan, Yufan Liu, Chunfeng Yuan, Bing Li, Weiming Hu, Yangxi Li, and
  Stephen Maybank.
\newblock Edp: An efficient decomposition and pruning scheme for convolutional
  neural network compression.
\newblock {\em IEEE Transactions on Neural Networks and Learning Systems},
  32(10):4499--4513, 2020.

\bibitem{ruder2016overview}
Sebastian Ruder.
\newblock An overview of gradient descent optimization algorithms.
\newblock {\em arXiv preprint arXiv:1609.04747}, 2016.

\bibitem{2020Knowledge}
R. Sadhukhan, A. Saha, J. Mukhopadhyay, and A. Patra.
\newblock Knowledge distillation inspired fine-tuning of tucker decomposed cnns
  and adversarial robustness analysis.
\newblock In {\em 2020 IEEE International Conference on Image Processing
  (ICIP)}, 2020.

\bibitem{samragh2019autorank}
Mohammad Samragh, Mojan Javaheripi, and Farinaz Koushanfar.
\newblock Autorank: Automated rank selection for effective neural network
  customization.
\newblock In {\em Proceedings of the ML-for-Systems Workshop at the 46th
  International Symposium on Computer Architecture (ISCA’19)}, 2019.

\bibitem{shi2019sparse}
Lei Shi, Xiaolin Huang, Yunlong Feng, and Johan Suykens.
\newblock Sparse kernel regression with coefficient-based lq-regularization.
\newblock {\em Journal of Machine Learning Research}, 20, 2019.

\bibitem{simonyan2014very}
Karen Simonyan and Andrew Zisserman.
\newblock Very deep convolutional networks for large-scale image recognition.
\newblock {\em arXiv preprint arXiv:1409.1556}, 2014.

\bibitem{sun2020deep}
Weize Sun, Shaowu Chen, Lei Huang, Hing~Cheung So, and Min Xie.
\newblock Deep convolutional neural network compression via coupled tensor
  decomposition.
\newblock {\em IEEE Journal of Selected Topics in Signal Processing},
  15(3):603--616, 2020.

\bibitem{swaminathan2020sparse}
Sridhar Swaminathan, Deepak Garg, Rajkumar Kannan, and Frederic Andres.
\newblock Sparse low rank factorization for deep neural network compression.
\newblock {\em Neurocomputing}, 398:185--196, 2020.

\bibitem{sze2017efficient}
Vivienne Sze, Yu-Hsin Chen, Tien-Ju Yang, and Joel~S Emer.
\newblock Efficient processing of deep neural networks: A tutorial and survey.
\newblock {\em Proceedings of the IEEE}, 105(12):2295--2329, 2017.

\bibitem{sze2020evaluate}
Vivienne Sze, Yu-Hsin Chen, Tien-Ju Yang, and Joel~S Emer.
\newblock How to evaluate deep neural network processors: Tops/w (alone)
  considered harmful.
\newblock {\em IEEE Solid-State Circuits Magazine}, 12(3):28--41, 2020.

\bibitem{tee2020information}
James Tee and Desmond~P Taylor.
\newblock Is information in the brain represented in continuous or discrete
  form?
\newblock {\em IEEE Transactions on Molecular, Biological and Multi-Scale
  Communications}, 6(3):199--209, 2020.

\bibitem{thakker2019compressing}
Urmish Thakker, Jesse Beu, Dibakar Gope, Chu Zhou, Igor Fedorov, Ganesh Dasika,
  and Matthew Mattina.
\newblock Compressing rnns for iot devices by 15-38x using kronecker products.
\newblock {\em arXiv preprint arXiv:1906.02876}, 2019.

\bibitem{tucker1963implications}
Ledyard~R Tucker.
\newblock Implications of factor analysis of three-way matrices for measurement
  of change.
\newblock {\em Problems in Measuring Change}, 15:122--137, 1963.

\bibitem{tucker1966some}
Ledyard~R Tucker.
\newblock Some mathematical notes on three-mode factor analysis.
\newblock {\em Psychometrika}, 31(3):279--311, 1966.

\bibitem{ullrich2017soft}
Karen Ullrich, Edward Meeds, and Max Welling.
\newblock Soft weight-sharing for neural network compression.
\newblock {\em arXiv preprint arXiv:1702.04008}, 2017.

\bibitem{vanrullen2003perception}
Rufin VanRullen and Christof Koch.
\newblock Is perception discrete or continuous?
\newblock {\em Trends in cognitive sciences}, 7(5):207--213, 2003.

\bibitem{varshney2006optimal}
Lav~R Varshney, Per~Jesper Sj{\"o}str{\"o}m, and Dmitri~B Chklovskii.
\newblock Optimal information storage in noisy synapses under resource
  constraints.
\newblock {\em Neuron}, 52(3):409--423, 2006.

\bibitem{vaswani2017attention}
Ashish Vaswani, Noam Shazeer, Niki Parmar, Jakob Uszkoreit, Llion Jones,
  Aidan~N Gomez, {\L}ukasz Kaiser, and Illia Polosukhin.
\newblock Attention is all you need.
\newblock {\em Advances in neural information processing systems}, 30, 2017.

\bibitem{veeramacheneni2022canonical}
Lokesh Veeramacheneni, Moritz Wolter, Reinhard Klein, and Jochen Garcke.
\newblock Canonical convolutional neural networks.
\newblock In {\em 2022 International Joint Conference on Neural Networks
  (IJCNN)}, pages 1--8. IEEE, 2022.

\bibitem{wang2023tensor}
Maolin Wang, Yu Pan, Xiangli Yang, Guangxi Li, and Zenglin Xu.
\newblock Tensor networks meet neural networks: A survey.
\newblock {\em arXiv preprint arXiv:2302.09019}, 2023.

\bibitem{wang2018wide}
Wenqi Wang, Yifan Sun, Brian Eriksson, Wenlin Wang, and Vaneet Aggarwal.
\newblock Wide compression: Tensor ring nets.
\newblock In {\em Proceedings of the IEEE Conference on Computer Vision and
  Pattern Recognition}, pages 9329--9338, 2018.

\bibitem{welch1984technique}
Terry~A. Welch.
\newblock A technique for high-performance data compression.
\newblock {\em Computer}, 17(06):8--19, 1984.

\bibitem{wen2020learning}
Weijing Wen, Fan Yang, Yangfeng Su, Dian Zhou, and Xuan Zeng.
\newblock Learning low-rank structured sparsity in recurrent neural networks.
\newblock In {\em 2020 IEEE International Symposium on Circuits and Systems
  (ISCAS)}, pages 1--4. IEEE, 2020.

\bibitem{wiedemann2020deepcabac}
Simon Wiedemann, Heiner Kirchhoffer, Stefan Matlage, Paul Haase, Arturo Marban,
  Talmaj Marin{\v{c}}, David Neumann, Tung Nguyen, Heiko Schwarz, Thomas
  Wiegand, et~al.
\newblock Deepcabac: A universal compression algorithm for deep neural
  networks.
\newblock {\em IEEE Journal of Selected Topics in Signal Processing},
  14(4):700--714, 2020.

\bibitem{wu2020hybrid}
Bijiao Wu, Dingheng Wang, Guangshe Zhao, Lei Deng, and Guoqi Li.
\newblock Hybrid tensor decomposition in neural network compression.
\newblock {\em Neural Networks}, 132:309--320, 2020.

\bibitem{wu2016quantized}
Jiaxiang Wu, Cong Leng, Yuhang Wang, Qinghao Hu, and Jian Cheng.
\newblock Quantized convolutional neural networks for mobile devices.
\newblock In {\em Proceedings of the IEEE conference on computer vision and
  pattern recognition}, pages 4820--4828, 2016.

\bibitem{4100845}
Peng Xu, Yin Tian, Huafu Chen, and Dezhong Yao.
\newblock Lp norm iterative sparse solution for eeg source localization.
\newblock {\em IEEE Transactions on Biomedical Engineering}, 54(3):400--409,
  2007.

\bibitem{xu2019trained}
Yuhui Xu, Yuxi Li, Shuai Zhang, Wei Wen, Botao Wang, Wenrui Dai, Yingyong Qi,
  Yiran Chen, Weiyao Lin, and Hongkai Xiong.
\newblock Trained rank pruning for efficient deep neural networks.
\newblock In {\em 2019 Fifth Workshop on Energy Efficient Machine Learning and
  Cognitive Computing-NeurIPS Edition (EMC2-NIPS)}, pages 14--17. IEEE, 2019.

\bibitem{yang2020learning}
Huanrui Yang, Minxue Tang, Wei Wen, Feng Yan, Daniel Hu, Ang Li, Hai Li, and
  Yiran Chen.
\newblock Learning low-rank deep neural networks via singular vector
  orthogonality regularization and singular value sparsification.
\newblock In {\em Proceedings of the IEEE/CVF conference on computer vision and
  pattern recognition workshops}, pages 678--679, 2020.

\bibitem{ye2020block}
Jinmian Ye, Guangxi Li, Di Chen, Haiqin Yang, Shandian Zhe, and Zenglin Xu.
\newblock Block-term tensor neural networks.
\newblock {\em Neural Networks}, 130:11--21, 2020.

\bibitem{yin2022batude}
Miao Yin, Huy Phan, Xiao Zang, Siyu Liao, and Bo Yuan.
\newblock Batude: Budget-aware neural network compression based on tucker
  decomposition.
\newblock In {\em Proceedings of the AAAI Conference on Artificial
  Intelligence}, 2022.

\bibitem{yin2021towards}
Miao Yin, Yang Sui, Siyu Liao, and Bo Yuan.
\newblock Towards efficient tensor decomposition-based dnn model compression
  with optimization framework.
\newblock In {\em Proceedings of the IEEE/CVF Conference on Computer Vision and
  Pattern Recognition}, pages 10674--10683, 2021.

\bibitem{yin2022hodec}
Miao Yin, Yang Sui, Wanzhao Yang, Xiao Zang, Yu Gong, and Bo Yuan.
\newblock Hodec: Towards efficient high-order decomposed convolutional neural
  networks.
\newblock In {\em Proceedings of the IEEE/CVF Conference on Computer Vision and
  Pattern Recognition}, pages 12299--12308, 2022.

\bibitem{zhang2018systematic}
Tianyun Zhang, Shaokai Ye, Kaiqi Zhang, Jian Tang, Wujie Wen, Makan Fardad, and
  Yanzhi Wang.
\newblock A systematic dnn weight pruning framework using alternating direction
  method of multipliers.
\newblock In {\em Proceedings of the European Conference on Computer Vision
  (ECCV)}, pages 184--199, 2018.

\bibitem{zhang2015accelerating}
Xiangyu Zhang, Jianhua Zou, Kaiming He, and Jian Sun.
\newblock Accelerating very deep convolutional networks for classification and
  detection.
\newblock {\em IEEE transactions on pattern analysis and machine intelligence},
  38(10):1943--1955, 2015.

\bibitem{zhao2021semi}
Hengling Zhao, Yipeng Liu, Xiaolin Huang, and Ce Zhu.
\newblock Semi-tensor product-based tensordecomposition for neural network
  compression.
\newblock {\em arXiv preprint arXiv:2109.15200}, 2021.

\bibitem{zhao2016tensor}
Qibin Zhao, Guoxu Zhou, Shengli Xie, Liqing Zhang, and Andrzej Cichocki.
\newblock Tensor ring decomposition.
\newblock {\em arXiv preprint arXiv:1606.05535}, 2016.

\bibitem{zimmer2022compression}
Max Zimmer, Christoph Spiegel, and Sebastian Pokutta.
\newblock Compression-aware training of neural networks using frank-wolfe.
\newblock {\em arXiv preprint arXiv:2205.11921}, 2022.

\bibitem{ziv1977universal}
Jacob Ziv and Abraham Lempel.
\newblock A universal algorithm for sequential data compression.
\newblock {\em IEEE Transactions on information theory}, 23(3):337--343, 1977.

\bibitem{ziv1978compression}
Jacob Ziv and Abraham Lempel.
\newblock Compression of individual sequences via variable-rate coding.
\newblock {\em IEEE transactions on Information Theory}, 24(5):530--536, 1978.

\bibitem{zoph2016neural}
Barret Zoph and Quoc~V Le.
\newblock Neural architecture search with reinforcement learning.
\newblock {\em arXiv preprint arXiv:1611.01578}, 2016.

\bibitem{zou2006sparse}
Hui Zou, Trevor Hastie, and Robert Tibshirani.
\newblock Sparse principal component analysis.
\newblock {\em Journal of computational and graphical statistics},
  15(2):265--286, 2006.

\end{thebibliography}
}
%




\end{document}